\documentclass[lettersize,journal]{IEEEtran}
\usepackage{amsmath,amsfonts}
\usepackage{algorithmic}
\usepackage{array}
\usepackage{textcomp}
\usepackage{stfloats}
\usepackage{url}
\usepackage{verbatim}
\usepackage{graphicx}
\usepackage{xcolor}

\newif\ifrevisionmarked
\revisionmarkedtrue 
\ifrevisionmarked
  \usepackage[draft,defaultcolor=black,addedmarkup=colored,deletedmarkup=sout,commandnameprefix=ifneeded]{changes}
  \definecolor{RevisionDeleted}{HTML}{000000}
  \setdeletedmarkup{\textcolor{RevisionDeleted}{\sout{#1}}}
\else
  \usepackage[final,commandnameprefix=ifneeded]{changes}
\fi
\definecolor{RevisionAdded}{HTML}{000000}
\setaddedmarkup{\textcolor{RevisionAdded}{#1}}

\definecolor{RevisionCurrent}{HTML}{000000}
\definecolor{SectionPendingRed}{HTML}{9B5C60}
\providecommand{\revnow}[1]{%
    \ifrevisionmarked
        \textcolor{RevisionCurrent}{#1}%
    \else
        #1%
    \fi
}

\usepackage{hyperref}
\usepackage{float}  
\usepackage{subcaption}

\def\BibTeX{{\rm B\kern-.05em{\sc i\kern-.025em b}\kern-.08em
    T\kern-.1667em\lower.7ex\hbox{E}\kern-.125emX}}
\usepackage{balance}
\begin{document}
\title{CRFCAN: A Complex-Valued Cross-Domain Residual Network for Joint Channel and Phase Noise Estimation in Sub-THz OFDM Systems}
\author{Ruilin~Wang and Xiaodai~Dong%
\thanks{The authors are with the Department of Electrical and Computer Engineering, University of Victoria, Victoria, BC V8P 5C2, Canada.}%
\thanks{Corresponding author: Xiaodai Dong (email: xdong@ece.uvic.ca).}%
}

\maketitle

\begin{abstract}
The coupling between multipath propagation and phase
noise (PN) poses challenges for joint channel and PN
estimation in sub-terahertz (sub-THz) cyclic-prefix OFDM
systems.
This paper proposes CRFCAN, a complex-valued residual
FFT convolutional attention network for single-input
single-output systems using comb-type pilots only.
Its shared backbone alternately refines time- and
frequency-domain features through complex-valued residual
channel attention and PN-inspired multiplicative updates,
producing channel and PN estimates in their respective
domains.
An unfolded PN refinement tail combines least-squares
updates with feature-guided complex-valued CNN corrections.
Simulation results show that CRFCAN achieves lower channel
estimation error and bit error rate than the evaluated
baselines in both training TDL-A/B/E + 3GPP RAN4 and TDL-C/D + 3GPP RAN1 generalization settings, together with lower PN estimation
error at medium-to-high signal-to-noise ratios.
Further tests without fine-tuning demonstrate its
applicability across different data modulations, OFDM
parameters, channel models, and PN spectra.
\end{abstract}


\section{INTRODUCTION}

\IEEEPARstart{T}{he} rapid proliferation of data-intensive
applications, such as ultra-high-definition video streaming
and immersive extended reality, has driven growing interest
in the sub-terahertz (sub-THz) band (100--300 GHz) as a key
enabling technology for future 6G wireless networks
\cite{itu2023imt2030}.
Operating in the sub-THz spectrum provides access to
multi-gigahertz bandwidths, offering the potential to support
terabit-per-second (Tbps) data rates
\cite{jiang2024terahertz}.
However, realizing this potential requires addressing
hardware impairments, including phase noise (PN) arising
from high-frequency local oscillators
\cite{chen2025review}.
In orthogonal frequency-division multiplexing (OFDM)
systems, PN can introduce common phase error (CPE) and
inter-carrier interference (ICI)
\cite{armada2001understanding}.
Approximating PN-induced ICI as additive white Gaussian
noise (AWGN) neglects its structured coupling across
subcarriers and can limit receiver performance when phase
fluctuations within an OFDM symbol are significant
\cite{zou2007compensation}.
These distortions, together with frequency-selective fading,
motivate joint estimation of channel state information
(CSI) and PN.

A range of model-based methods have been developed to address
the combined effects of channel distortion and PN.
One approach combines pilot-aided channel estimation with
phase-tracking reference signals (PTRS) for phase
compensation.
Basic PTRS processing primarily targets CPE, with
reference-signal density balancing tracking accuracy against
pilot overhead \cite{qi2018ptrs}.
For higher-frequency operation, enhanced block-PTRS designs
also enable the estimation and compensation of PN-induced
ICI \cite{levanen2021mobile}.
\textcolor{RevisionCurrent}{%
Classical ICI compensation also uses a reduced set of
Fourier coefficients to represent the PN phasor and
construct a frequency-domain compensation filter
\cite{petrovic2007effects}.%
}
Other methods explicitly formulate joint channel and PN
estimation.
Zou et al.\ \cite{zou2007compensation} employ alternating
least-squares (LS) updates to jointly estimate the channel
and PN using block-type pilots, followed by iterative data
and PN estimation using comb-type pilots.
Salim et al.\ \cite{salim2014channel} develop an expectation
conditional maximization (ECM) framework for joint estimation
of the channel, PN, and carrier frequency offset (CFO) in
OFDM systems.
To improve computational efficiency,
Wang et al.\ \cite{wang2017effective} exploit the unit-modulus
structure of the PN phasor and develop low-complexity joint
estimators using majorization--minimization, dimensionality
reduction, and regularization.
For time-varying channels, Linsalata and Ksairi
\cite{linsalata2023joint} use separate basis expansion models
(BEMs) to represent channel and PN variations across
multiple OFDM symbols and derive an approximate
maximum-likelihood method for jointly estimating their
coefficients.
In massive-MIMO downlink systems,
Zhang et al.\ \cite{zhang2020downlink} model PN as a
correlated perturbation to the pilot matrix and propose
PN-aware sparse Bayesian learning (PNA-SBL) for synchronous
PN, with iterative updates of channel estimates, PN-related
variables, and hyperparameters.
For hybrid near- and far-field THz ultra-massive MIMO
(UM-MIMO) channels, Li and Madhukumar combine
sparsifying-matrix learning with Bayesian channel
estimation~\cite{li2025hybrid}.

\textcolor{RevisionCurrent}{%
Learning-based receivers have been investigated for PN
compensation, joint estimation, and data detection under
a range of signal models.%
}
Neshaastegaran and Jian
\cite{neshaastegaran2024deep} address high-order modulation
over AWGN backhaul channels with random-walk PN, using a
convolutional neural network (CNN) to process
phase-likelihood grids constructed from pilot-assisted
initial estimates.
Seo et al.\ \cite{seo2025deep} combine a bidirectional long
short-term memory (Bi-LSTM) network with a feedforward
neural network (FNN) for sub-THz OFDM reception with
available CSI, processing received signals and CSI from
neighboring subcarriers to exploit frequency-domain
dependencies and implicitly compensate for PN during data
detection.
Marasinghe et al.\ \cite{marasinghe2025phase} develop
DeepSRX for single-carrier links with flat Rician fading and
transmitter and receiver PN, combining pilot-derived latent
channel--PN features with received data in a convolutional
residual network that directly produces
\textcolor{RevisionCurrent}{bit log-likelihood ratios (LLRs)}
rather than explicit channel and PN estimates.
Assuming a frequency-selective channel that remains constant
within each packet, Mohammadian et al.\
\cite{mohammadian2021deep} employ block- and comb-type pilots
with two fully connected deep neural networks (DNNs) that
refine initial channel and PN estimates, followed by a
residual CNN that denoises the initially detected data
symbols.
For doubly selective OFDM channels, Mattu and Chockalingam
\cite{mattu2022learning} combine two-dimensional CNN-based
channel estimation with model-based PN estimation and
compensation, then reuse the same CNN to refine the channel
estimate from the compensated observations.
\textcolor{RevisionCurrent}{%
Model-driven learning has also been applied to PN-impaired
OFDM reception: Li et al.\ \cite{li2024intelligent}
develop a semi-blind receiver for joint channel estimation,
PN compensation, and data detection by unfolding an
optimization algorithm based on the alternating direction
method of multipliers and minorization--maximization,
under constant-modulus PN and discrete-symbol constraints.%
}

\textcolor{RevisionCurrent}{%
Learning-based channel estimation has also been investigated
for wideband hybrid-field THz UM-MIMO
channels~\cite{li2025modeldriven}, including a
dual-attention-aided network obtained by unfolding
Bayesian estimation~\cite{li2026wideband}.%
}
\textcolor{RevisionCurrent}{%
Related work explores complex-valued and time--frequency
feature processing for channel estimation and prediction
without explicitly modeling oscillator PN.
Jiang et al.~\cite{jiang2021dual} connect CNN estimators
in the spatial--frequency and angle--delay domains through
Fourier transforms to refine pilot-based LS channel estimates.
Lee et al.~\cite{lee2025alternative} extend dual-domain
estimation to spatial--frequency--time and
angle--delay--Doppler representations, with selective
meta-learning in the latter domain to facilitate adaptation
to changing channels.
Leng et al.\ \cite{leng2025unveiling} use complex-valued
Transformers to reconstruct the full time--frequency
channel response from pilot-based LS estimates.%
}
For MIMO CSI prediction, Ren and Wang's CMTF-Net
\cite{ren2026cmtf} combines learnable multi-scale short-time
Fourier representations, complex-valued attention, and
bidirectional time--frequency feature interaction to predict
future CSI from historical observations.

\textcolor{RevisionCurrent}{%
However, the learning-based approaches reviewed above do not
combine shared time--frequency feature learning with explicit
channel and PN outputs within a single jointly trained
network for sub-THz single-input single-output (SISO)
cyclic-prefix OFDM (CP-OFDM) reception using only comb-type
pilots.%
}

\textcolor{RevisionCurrent}{%
In this setting, an estimator can exploit two sources
of information: the statistical structure of the channel
and PN learned from training data, and the known pilots
that constrain their estimates through the received
observations in the current frame.
The proposed design combines these sources using a shared
backbone that processes frequency- and time-domain features
and connects to two estimation tails for the channel
and PN, respectively.
Supervision of both outputs allows the two estimation tasks
to shape the shared features, while the resulting estimates
support subsequent channel equalization and PN compensation.%
}

\textcolor{RevisionCurrent}{%
Based on these considerations, we propose CRFCAN
(complex-valued residual FFT convolutional attention
network) for pilot-aided joint channel and PN estimation
in sub-THz SISO CP-OFDM systems.%
}
The main contributions are summarized as follows:

\begin{itemize}
    \item \textbf{%
    Jointly Supervised Cross-Domain Residual Architecture%
    }:
    We develop a shared complex-valued residual backbone
    with repeated FFT/IFFT transitions and feature
    refinement in both domains.
    \textcolor{RevisionCurrent}{%
The network produces channel estimates in the frequency
domain and PN estimates in the time domain,
with joint supervision of the shared cross-domain
representation.%
}

    \item \textbf{%
    Complex-Valued Residual Attention and PN-Inspired
    Rotation%
    }:
    \textcolor{RevisionCurrent}{%
We extend the residual channel attention block
(RCAB)~\cite{zhang2018image} to complex-valued processing,
yielding the complex residual channel attention block
(CRCAB).%
}
    We further introduce a phase-rotation residual channel
    attention block (PRCAB), which converts predicted
    feature-dependent phase angles into complex phasors for
    gated residual updates.
    \textcolor{RevisionCurrent}{%
    Together, these blocks combine complex-valued feature
    refinement with PN-inspired multiplicative updates.
    The selected combination retains similar high-SNR
    performance to an all-CRCAB backbone with fewer
    parameters.%
    }

    \item \textbf{\textcolor{RevisionCurrent}{%
Unfolded PN Refinement Tail%
}}:
\textcolor{RevisionCurrent}{%
Using the received signal, known pilots, and channel estimate,
we construct a fixed-depth unfolded tail that alternates
regularized LS updates with complex-valued CNN corrections
guided by shared features.%
}
\end{itemize}

The remainder of this paper is organized as follows.
Section II presents the system model and formulates the
joint channel and PN estimation problem.
Section III describes the CRFCAN architecture and its
constituent modules.
\textcolor{RevisionCurrent}{%
Section IV specifies the experimental setup.
Section V presents the performance comparisons,
generalization tests, ablation studies, and
computational-cost evaluation.
Finally, Section VI concludes the paper.%
}

\begin{figure*}[!t]
    \centering
    \includegraphics[width=\textwidth]{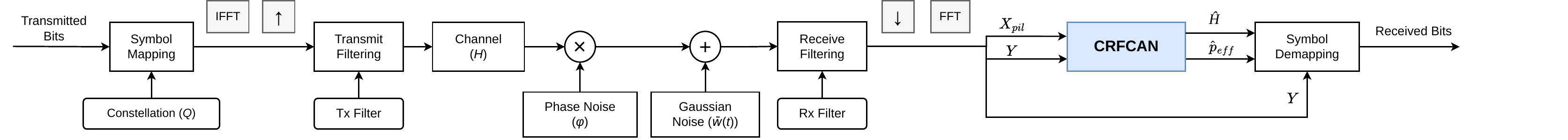}
    \caption{\textcolor{RevisionCurrent}{%
    System model of the considered sub-THz OFDM link
    with channel and PN impairments.%
    }}
    \label{fig:system_model}
\end{figure*}

\section{System Model}
\label{sec:system_model}
\added{%
As illustrated in Fig.~\ref{fig:system_model}, we consider a pilot-aided sub-THz SISO CP-OFDM system for joint channel and PN estimation.
The signal model accounts for frequency-selective, time-varying multipath propagation and oscillator PN, together with transmit pulse shaping and receive matched filtering.
The corresponding signal processing chain is described as follows.%
} 

\subsection{OFDM System Model}
\label{subsec:ofdm}

\begingroup
\ifrevisionmarked
    \color{black}
\fi

As shown in Fig.~\ref{fig:system_model}, information bits are
mapped to complex-valued data symbols and multiplexed with
known pilots.
Let $X_m[k]$ denote the data or pilot symbol transmitted on
subcarrier $k$ of OFDM symbol $m$, where
$k=0,\ldots,N_c-1$ and $m\in\mathbb{Z}$ indexes the continuous
symbol stream.
The receiver observes a block comprising symbols
$m=0,\ldots,N_t-1$.
An $N_c$-point IFFT generates the corresponding time-domain
samples:
\begin{equation}
    x_m[n]
    =
    \frac{1}{N_c}
    \sum_{k=0}^{N_c-1}
    X_m[k]e^{j2\pi kn/N_c},
    \quad n=0,\ldots,N_c-1.
    \label{eq:ofdm_tx_ifft}
\end{equation}

A CP of $L_{\mathrm{CP}}$ samples is inserted before each useful
symbol interval.
The CP-extended symbols are then serialized and pulse-shaped
by a unit-energy root-Nyquist transmit pulse
$g_{\mathrm{tx}}(t)$, normalized such that
$\int_{-\infty}^{\infty}|g_{\mathrm{tx}}(t)|^2\,dt=1$.
Define the sampling interval $T_s=1/(N_c\Delta f)$ and the
total symbol duration
$T_{\mathrm{sym}}=(N_c+L_{\mathrm{CP}})T_s$, where $\Delta f$
is the subcarrier spacing.
With the time origin at the start of the first observed
useful symbol interval, the transmitted baseband waveform is
\begin{equation}
    x(t)
    =
    \sum_{m\in\mathbb{Z}}
    \sum_{n=-L_{\mathrm{CP}}}^{N_c-1}
    x_m[(n)_{N_c}]
    g_{\mathrm{tx}}\!\left(t-mT_{\mathrm{sym}}-nT_s\right),
    \label{eq:ofdm_tx_waveform}
\end{equation}
where $(\cdot)_{N_c}$ denotes modulo-$N_c$ indexing.
This expression includes the preceding and following symbols
in addition to the observed block.

For a time-varying multipath channel $h(\tau,t)$, the received
signal before matched filtering is
\begin{equation}
    r(t)
    =
    e^{j\phi(t)}
    \int_{-\infty}^{\infty}
    h(\tau,t)x(t-\tau)\,d\tau
    +
    \bar{w}(t),
    \label{eq:ofdm_rx_waveform}
\end{equation}
where $\phi(t)$ represents oscillator PN, with only
receiver-side PN considered, and $\bar{w}(t)$ is zero-mean,
circularly symmetric complex AWGN with two-sided power
spectral density $N_0$, corresponding to $N_0/2$ in each
real and imaginary component.
Ideal symbol timing and no residual carrier-frequency offset
are assumed.

The receiver applies the matched filter
$g_{\mathrm{rx}}(t)=g_{\mathrm{tx}}^{*}(-t)$, followed by
sampling and CP removal.
Here, ${}^{*}$ denotes complex conjugation, and $\star$
denotes convolution.
After compensating for the fixed transmit- and
receive-filter delays, we set $t=0$ at the first sample
of the first observed FFT window.
The FFT window for OFDM symbol $m$ therefore starts
at $t_m^{\mathrm{FFT}}=mT_{\mathrm{sym}}$.
The $N_c$ samples supplied to the FFT are
\begin{equation}
    y_m[n]
    =
    (r\star g_{\mathrm{rx}})
    \bigl(t_m^{\mathrm{FFT}}+nT_s\bigr),
    \quad n=0,\ldots,N_c-1.
    \label{eq:ofdm_rx_samples}
\end{equation}
An unscaled $N_c$-point FFT transforms the retained samples
into frequency-domain observations:
\begin{equation}
    Y_m[k]
    =
    \sum_{n=0}^{N_c-1}
    y_m[n]e^{-j2\pi kn/N_c},
    \quad k=0,\ldots,N_c-1.
    \label{eq:ofdm_rx_fft}
\end{equation}
The finite-length square-root raised-cosine (SRRC)
implementation and numerical noise calibration are
specified in Section~IV.

The received observations form a grid
$\mathbf{Y}\in\mathbb{C}^{N_c\times N_t}$, with
$[\mathbf{Y}]_{k,m}=Y_m[k]$, where rows correspond to
subcarriers and columns to OFDM symbols.
The transmitted pilot symbols and their locations are known
to the receiver, whereas the transmitted data symbols are
unknown.

\par
\endgroup

\subsection{Sub-THz Channel Model}
\label{subsec:channel_model}

\begingroup
\ifrevisionmarked
    \color{black}
\fi

The SISO propagation channel is modeled as a time-varying
tapped delay line (TDL) using the profiles specified in
3GPP TR~38.901 for carrier frequencies from 0.5 to
100~GHz \cite{3gpp38901}.
Its continuous-time channel impulse response is
\begin{equation}
    h(\tau,t)
    =
    \sum_{\ell=0}^{L_{\mathrm{tap}}-1}
    a_\ell(t)\delta(\tau-\tau_\ell),
    \label{eq:tdl_continuous_channel}
\end{equation}
where $L_{\mathrm{tap}}$ is the number of taps,
$\tau_\ell$ is the propagation delay of tap $\ell$,
$a_\ell(t)$ is its time-varying complex gain, and
$\delta(\cdot)$ denotes the Dirac delta function.
The tap delays and average powers remain fixed over an
observed block, while the complex gains evolve with time.

The selected TDL profile specifies the relative tap delays
and average powers.
For LOS profiles, the diffuse component powers are rescaled
according to the prescribed channel-wide K-factor
$K_{\mathrm{ch}}$, defined as the ratio of the specular
power to the total diffuse power across all taps.
The first-tap Rician factor is determined by the specular
and diffuse powers within that tap.
After the K-factor adjustment, the tap delays are rescaled
to give unit RMS delay spread under the adjusted tap-power
weights, following Section~7.7.6 of \cite{3gpp38901}.
Let $\widetilde{\tau}_\ell$ denote the resulting dimensionless
tap delay.
For a target RMS delay spread $\sigma_\tau$, the physical
tap delay is
\begin{equation}
    \tau_\ell
    =
    \sigma_\tau\widetilde{\tau}_\ell.
    \label{eq:tdl_delay_scaling}
\end{equation}
The average tap powers, denoted by
$\Omega_\ell=\mathbb{E}[|a_\ell(t)|^2]$, are normalized
such that
$\sum_{\ell=0}^{L_{\mathrm{tap}}-1}\Omega_\ell=1$.

For NLOS profiles, the tap-gain magnitudes are Rayleigh
distributed.
For LOS profiles, the first tap combines a specular
component with diffuse fading, giving a Rician-distributed
magnitude, while the remaining tap magnitudes are Rayleigh
distributed.
The diffuse fading processes are generated independently
across taps using a sum-of-sinusoids approximation to the
classical Jakes Doppler spectrum \cite{jakes1974microwave}.
The maximum Doppler frequency is
\begin{equation}
    f_d
    =
    \frac{vf_c}{c},
    \label{eq:tdl_max_doppler}
\end{equation}
where $v$ is the terminal speed, $f_c$ is the carrier
frequency, and $c$ is the speed of light.
The LOS specular component undergoes a deterministic phase
evolution with Doppler shift $f_{\mathrm{LOS}}=0.7f_d$,
as specified in Section~7.7.2 of \cite{3gpp38901}.

The selected profiles, scenario-dependent delay-spread
and K-factor statistics, the mobility distribution, and
the channel-sampling procedure used in the simulations
are specified in Section~IV.

\par
\endgroup

\subsection{Phase Noise Model}
\label{subsec:pn_model}

\begingroup
\ifrevisionmarked
    \color{black}
\fi

The receiver oscillator phase $\phi(t)$ is modeled as a
real-valued, zero-mean stationary Gaussian process,
independent of the propagation channel and transmitted
symbols.
Its prescribed single-sideband (SSB) phase-noise spectrum
follows the fractional pole--zero form of the phase-locked-loop
(PLL) model in 3GPP TR~38.803~\cite{3gpp38803}.
We refer to this model as RAN4, and to the two PN
parameter sets in TR~38.808~\cite{3gpp38808} as
RAN1 Set1 and Set2.
For a positive frequency offset $f$ from the carrier, the
spectral level in dBc/Hz is expressed as
\begin{equation}
    \begin{aligned}
        L(f;f_c)
        ={}& L_0(f_c)+\Delta_{\mathrm{cal}} \\
        &+
        10\log_{10}
        \left(
        \frac{
            \prod_{i=1}^{J_z}
            \left[1+(f/f_{z,i})^{\alpha_{z,i}}\right]
        }{
            \prod_{i=1}^{J_p}
            \left[1+(f/f_{p,i})^{\alpha_{p,i}}\right]
        }
        \right),
    \end{aligned}
    \label{eq:pn_psd_linear}
\end{equation}
where $L_0(f_c)$ is the spectral level parameter and
$\Delta_{\mathrm{cal}}$ is an additive PSD level adjustment.
The quantities $J_z$ and $J_p$ denote the numbers of zero
and pole terms, respectively.
The corner frequencies $f_{z,i}$ and $f_{p,i}$, together
with the exponents $\alpha_{z,i}$ and $\alpha_{p,i}$,
determine the spectral shape.
Their dependence on $f_c$ is omitted from the notation
for brevity.

The reference spectra are specified at lower carrier
frequencies and are extended to the sub-THz operating
frequency through carrier-dependent scaling and spectral
fitting.
For a carrier frequency without a tabulated parameter set,
the spectral level parameter follows the ideal scaling
$L_0(f_c)=L_0(f_{\mathrm{base}})
+20\log_{10}(f_c/f_{\mathrm{base}})$,
where $f_{\mathrm{base}}$ is the reference carrier frequency.
The corner frequencies are then fitted at a frequency offset
$f_{\mathrm{fit}}$ to satisfy
\begin{equation}
    \begin{aligned}
        L(f_{\mathrm{fit}};f_c)
        ={}& L(f_{\mathrm{fit}};f_{\mathrm{base}}) \\
        &+20\log_{10}\!\left(
            \frac{f_c}{f_{\mathrm{base}}}
        \right)
        +\Delta_{\mathrm{FoM}}(f_c),
    \end{aligned}
    \label{eq:pn_carrier_scaling}
\end{equation}
with the same $\Delta_{\mathrm{cal}}$ for both spectra.
The $20\log_{10}(f_c/f_{\mathrm{base}})$ term represents
the phase-noise increase associated with ideal frequency
multiplication, while the additional degradation is described
by the phenomenological figure-of-merit (FoM) correction
\begin{equation}
    \Delta_{\mathrm{FoM}}(f_c)
    =
    \kappa_{\mathrm{FoM}}
    \log_{10}\!\left(
        \frac{f_c}{f_{\mathrm{base}}}
    \right),
    \label{eq:pn_fom_correction}
\end{equation}
where $\kappa_{\mathrm{FoM}}$ controls the additional
degradation per frequency decade.
\revnow{%
The carrier-scaling and spectral-fitting procedure follows
3GPP TR~38.803~\cite[Sec.~6.1.10.1]{3gpp38803}.%
}
Finally, $\Delta_{\mathrm{cal}}$ shifts the fitted spectrum
uniformly.

Under the small-phase SSB convention, the corresponding
target two-sided phase PSD is
\begin{equation}
    S_{\phi,\mathrm{tar}}^{(2)}(f)
    =
    10^{L(|f|;f_c)/10},
    \label{eq:pn_target_twosided_psd}
\end{equation}
in $\mathrm{rad}^2/\mathrm{Hz}$.
A single PN realization spans consecutive OFDM symbols.
\revnow{%
The reference carrier, FoM coefficient, PSD level adjustment,
modeled frequency range, low-frequency continuation, and
numerical synthesis settings are specified in
Section~\ref{subsec:sim_setup}.%
}

\par
\endgroup

\begin{figure*}[!t]
    \centering
    \includegraphics[width=\textwidth]{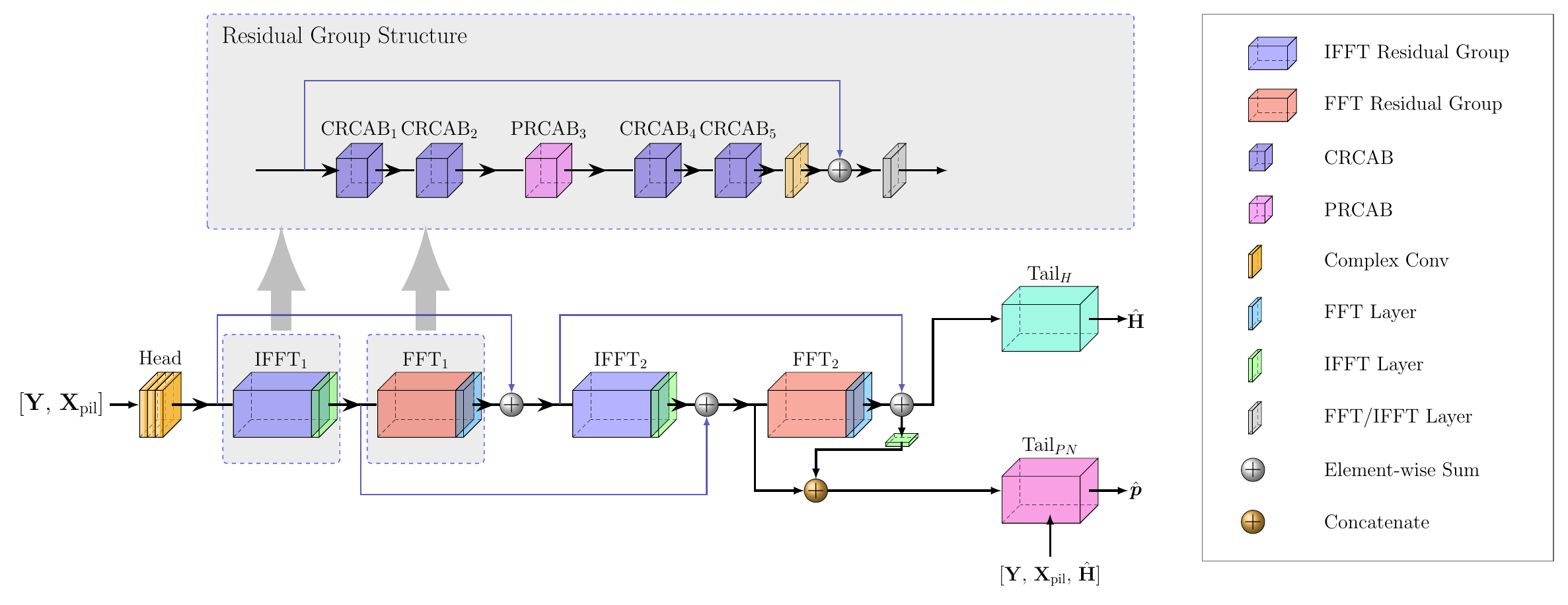} 
    \caption{Overall architecture of the proposed CRFCAN, consisting of a complex-convolution front-end, alternating IFFT/FFT residual groups for cross-domain refinement, and a dual-output tail for joint CSI and PN estimation.}
    \label{fig:CRFCAN_overview}
\end{figure*}

\subsection{Problem Formulation and Identifiability}
\label{subsec:problem_formulation}

\revnow{%
The supervision targets are a symbol-centre channel
reference and PN coefficients defined from the filtered
waveform.
Under the timing convention of
Section~\ref{subsec:ofdm}, let
$t_m^c=t_m^{\mathrm{FFT}}+(N_c-1)T_s/2$
denote the midpoint of the $N_c$ FFT sample times and
$g_{\ell,\mathrm{eff}}
=g_{\mathrm{tx}}\star d_\ell\star g_{\mathrm{rx}}$
the effective response of path $\ell$.
Here, $d_\ell$ implements the path delay $\tau_\ell$;
for an ideal delay, $d_\ell(t)=\delta(t-\tau_\ell)$.
Freezing the path gains at $t_m^c$, the circular
channel reference is%
}
\begin{equation}
    h_m^{\mathrm{ref}}[r]
    =
    \sum_{i\in\mathbb{Z}}
    \sum_{\ell=0}^{L_{\mathrm{tap}}-1}
    a_\ell(t_m^c)
    g_{\ell,\mathrm{eff}}
    \bigl((r+iN_c)T_s\bigr),
    \label{eq:channel_reference_cir}
\end{equation}
\revnow{%
for $r=0,\ldots,N_c-1$.
For each $r$, the sum combines all sampled lags $r+iN_c$
that share the same circular index.
Its unscaled DFT,
$H_m[k]=\operatorname{DFT}\{h_m^{\mathrm{ref}}\}[k]$,
defines the channel target.
The received waveform is generated using the time-varying
path gains in Section~II-B.%
}

\revnow{%
Let $\bar y_m[n]$ and $\bar y_m^\phi[n]$
denote noiseless received samples without and with PN,
using the same transmitted waveform, channel realization,
and receive processing.
The PN target is%
}
\begin{equation}
    p_{\mathrm{eff},m}[n]
    =
    \frac{\bar y_m^\phi[n]}{\bar y_m[n]},
    \qquad \bar y_m[n]\neq0.
    \label{eq:effective_pn_target}
\end{equation}
\revnow{%
PN multiplication and receive filtering do not generally
commute~\cite{piemontese2024discrete}.
For an isolated pulse without multipath, define
$g_{\mathrm{eff}}=g_{\mathrm{tx}}\star g_{\mathrm{rx}}$
and
$g_{\mathrm{eff}}^\phi
=(g_{\mathrm{tx}}e^{j\phi})\star g_{\mathrm{rx}}$.
Matched filtering gives%
}
\begin{equation}
    \frac{g_{\mathrm{eff}}^\phi(0)}
         {g_{\mathrm{eff}}(0)}
    =
    \frac{
        \displaystyle\int_{-\infty}^{\infty}
        |g_{\mathrm{tx}}(u)|^2e^{j\phi(u)}\,du
    }{
        \displaystyle\int_{-\infty}^{\infty}
        |g_{\mathrm{tx}}(u)|^2\,du
    }.
    \label{eq:pn_weighted_phasor}
\end{equation}
\revnow{%
This weighted phasor average can have a magnitude below one.
For the full OFDM waveform, interference among pulse and
multipath contributions can make the PN-free denominator
small, allowing the ratio magnitude to exceed one.%
}

\revnow{%
The circular reference output and received samples satisfy%
}
\begingroup
\ifrevisionmarked
    \color{RevisionCurrent}
\fi
\begin{equation}
    \begin{aligned}
        z_m[n]
        &=
        \sum_{r=0}^{N_c-1}
        h_m^{\mathrm{ref}}[r]
        x_m[(n-r)_{N_c}],\\
        y_m[n]
        &=
        p_{\mathrm{eff},m}[n]z_m[n]
        +e_m[n]+w_m[n],\\
        e_m[n]
        &=
        p_{\mathrm{eff},m}[n]
        \bigl(\bar y_m[n]-z_m[n]\bigr).
    \end{aligned}
    \label{eq:joint_time_domain_model}
\end{equation}
\endgroup
\revnow{%
The last identity holds for $\bar y_m[n]\neq0$.
Here, $w_m[n]$ is the filtered additive noise.
The difference $\bar y_m[n]-z_m[n]$ measures the mismatch
between the PN-free received samples and the circular
channel reference, including within-symbol channel variation.
Multiplication by $p_{\mathrm{eff},m}[n]$ gives the
received-signal residual $e_m[n]$.
Numerical safeguards and channel-memory acceptance are
specified in Section~\ref{sec:experimental_setup}.%
}

\revnow{%
Define
$P_m[k]=N_c^{-1}\operatorname{DFT}
\{p_{\mathrm{eff},m}\}[k]$.
The frequency-domain observations are%
}
\begin{equation}
    \begin{aligned}
        Y_m[k]
        ={}&
        \sum_{q=0}^{N_c-1}
        P_m[(k-q)_{N_c}]H_m[q]X_m[q]\\
        &+E_m[k]+W_m[k],
    \end{aligned}
    \label{eq:joint_frequency_domain_model}
\end{equation}
\revnow{%
where $E_m[k]$ and $W_m[k]$ are the unscaled DFTs
of the residual and noise.
The coefficient $P_m[0]$ multiplies every subcarrier within
symbol $m$, with its phase representing CPE.
The terms with $q\neq k$ describe PN-induced intercarrier
coupling.%
}

\revnow{%
Define $\mathbf{H},\mathbf{p}_{\mathrm{eff}}
\in\mathbb{C}^{N_c\times N_t}$ by
$[\mathbf{H}]_{k,m}=H_m[k]$ and
$[\mathbf{p}_{\mathrm{eff}}]_{n,m}=p_{\mathrm{eff},m}[n]$.
Their rows index subcarriers and time-domain samples,
respectively, and their columns index OFDM symbols.
Let $\mathbf{X}_{\mathrm{pil}}$ contain known pilots
and zeros elsewhere.
CRFCAN estimates%
}
\begin{equation}
    \bigl(
        \widehat{\mathbf{H}},
        \widehat{\mathbf{p}}_{\mathrm{eff}}
    \bigr)
    =
    f_\theta(\mathbf{Y},\mathbf{X}_{\mathrm{pil}}),
    \label{eq:joint_estimation_mapping}
\end{equation}
\revnow{%
where $\theta$ denotes the network parameters.
Under the prescribed training distribution, the estimates
minimizing the mean squared error (MSE) are the conditional
means of the channel and PN
targets given $(\mathbf{Y},\mathbf{X}_{\mathrm{pil}})$.
}

\revnow{%
The summation term in
\eqref{eq:joint_frequency_domain_model} is invariant under%
}
\begin{equation}
    P_m[k]\longmapsto\kappa_mP_m[k],
    \qquad
    H_m[k]\longmapsto H_m[k]/\kappa_m,
    \label{eq:joint_scaling_ambiguity}
\end{equation}
\revnow{%
for nonzero complex $\kappa_m$, with the residual
held fixed.
If the nonzero channel reference is constrained to be
identical across symbols, the same scaling factor must
be used for every symbol.
Pilots alone cannot remove this scaling ambiguity;
estimation of the separate targets also relies on channel
and PN statistics, including cross-symbol correlation
and PN magnitude statistics.%
}

\section{Proposed Network}
\subsection{Network Architecture and Cross-Domain Residual Groups}
\label{subsec:network_architecture}
\label{subsec:cross_domain_rg}

\begin{figure}[!t]
    \centering
    \includegraphics[width=\columnwidth]{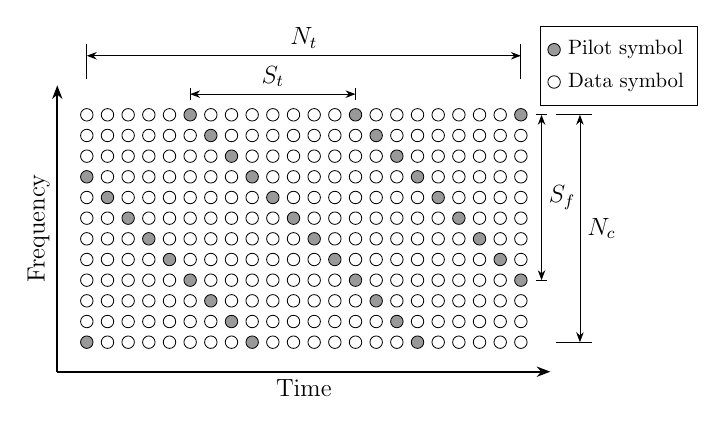}
    \caption{Comb-type pilot pattern with cyclic shift
    across consecutive OFDM symbols. $S_f$ denotes the pilot
    spacing in subcarriers, and $S_t$ denotes the pilot
    repetition interval in OFDM symbols on a given subcarrier.}
    \label{fig:pilot_design}
\end{figure}

\revnow{%
As shown in Fig.~\ref{fig:CRFCAN_overview}, CRFCAN
comprises a convolutional front-end, a cross-domain
residual backbone, and channel and PN estimation tails.
The input contains two complex-valued channels,
$\mathbf{Y}$ and $\mathbf{X}_{\mathrm{pil}}$, forming
a $2\times N_c\times N_t$ tensor with zeros at
non-pilot positions in $\mathbf{X}_{\mathrm{pil}}$.
The comb-pilot offset shifts cyclically across
OFDM symbols~\cite{coleri2002channel}, as illustrated
in Fig.~\ref{fig:pilot_design}; its configuration
is specified in Section~\ref{sec:experimental_setup}.
Feature processing uses standard complex
convolutions~\cite{trabelsi2017deep,lee2022complex}
and the split activation
$\operatorname{CReLU}(z)
=\operatorname{ReLU}(\Re\{z\})
+j\operatorname{ReLU}(\Im\{z\})$.%
}

\revnow{%
In the reference model of
Section~\ref{subsec:problem_formulation},
the channel acts multiplicatively in frequency
and through circular convolution in time,
whereas PN acts multiplicatively in time and
through circular convolution in frequency.
The backbone alternates feature processing between
these complementary domains.
Following the residual-group organization of
RCAN~\cite{zhang2018image}, each group contains
residual attention blocks and a group-level
skip connection~\cite{he2016deep}.%
}

\revnow{%
Let $\mathbf{F}_0\in
\mathbb{C}^{C\times N_c\times N_t}$
denote the frequency-domain features produced
by the front-end, where $C$ is the feature-channel
count.
The backbone contains an even number $G$ of
alternating IFFT and FFT residual groups.
For group $g$, $\mathcal{H}_g$ consists of
the block sequence followed by a complex convolution.
Its output is added to the group input and
transformed as%
}
\begin{equation}
    \mathbf{F}'_g
    =
    \mathcal{T}_g\!\left(
        \mathbf{F}_{g-1}
        +
        \mathcal{H}_g(\mathbf{F}_{g-1})
    \right),
    \label{eq:RG}
\end{equation}
\revnow{%
where $\mathcal{T}_g$ is an IFFT for odd $g$
and an FFT for even $g$.
An IFFT-RG processes frequency-domain features before
applying the IFFT, whereas an FFT-RG processes time-domain
features before applying the FFT.
The transforms act along the length-$N_c$ axis,
independently for each feature channel and
OFDM symbol, using unitary normalization
$1/\sqrt{N_c}$ in both directions.
This internal convention differs from the
waveform transform pair in
Section~\ref{subsec:ofdm}.%
}

\revnow{%
The first group establishes the time-domain
representation, while subsequent groups fuse
their transformed outputs with the most recent
features in the same domain:%
}
\begin{equation}
    \begin{aligned}
        \mathbf{F}_1
        &=
        \mathbf{F}'_1,\\
        \mathbf{F}_g
        &=
        \mathbf{F}_{g-2}
        +
        2\,\operatorname{sigmoid}(\gamma_g)
        \mathbf{F}'_g,
        \quad g=2,\ldots,G.
    \end{aligned}
    \label{eq:rg_domain_fusion}
\end{equation}
\revnow{%
Here, $\gamma_g$ is a learned real scalar
initialized to zero.
The intra-group skip precedes the transform,
whereas the gated skip connects successive
representations in the same domain.
Each group contains one PRCAB between CRCABs,
whose operations are described in
Section~\ref{subsec:residual_attention_blocks}.
The channel tail reads $\mathbf{F}_G$,
and the PN tail uses $\mathbf{F}_{G-1}$
together with $\mathcal{IFFT}(\mathbf{F}_G)$.
The group and block arrangements and counts
are specified in Section~\ref{sec:experimental_setup}.%
}
\subsection{Residual Attention Blocks for Joint Modeling}
\label{subsec:residual_attention_blocks}

\begingroup
\ifrevisionmarked
    \setaddedmarkup{\textcolor{black}{#1}}
\fi

\subsubsection{CRCAB}

\begin{figure}[!t]
    \centering
    \includegraphics[width=0.5\textwidth]{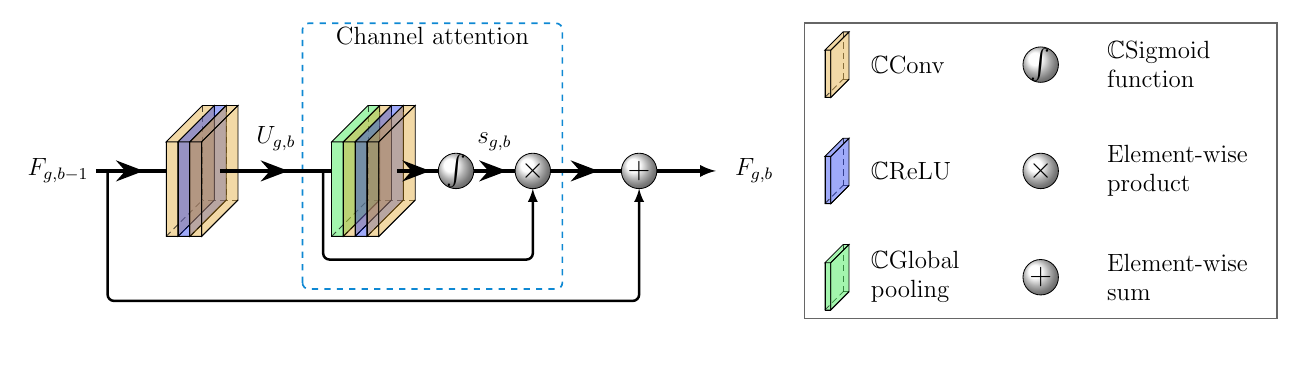}
    \caption{Complex Residual Channel Attention Block (CRCAB).}
    \label{fig:CRCAB}
\end{figure}

We first introduce the complex residual channel attention
block (CRCAB), which extends the residual channel attention
design to complex-valued feature maps.
Given the input feature
$\mathbf{F}_{g,b-1}\in\mathbb{C}^{C\times N_c\times N_t}$,
CRCAB learns an intermediate feature
$\mathbf{U}_{g,b}\in\mathbb{C}^{C\times N_c\times N_t}$
through two stacked complex convolutions
\added{with $\operatorname{CReLU}$ between them},
and applies channel-wise attention to
\added{reweight}
$\mathbf{U}_{g,b}$ before the residual addition,
as illustrated in Fig.~\ref{fig:CRCAB}.
\added{Through local complex-valued convolutions and channel
attention, CRCAB learns and reweights features of the coupled
channel and PN structure in
\eqref{eq:joint_frequency_domain_model}.}
\added{The first block receives
$\mathbf{F}_{g,0}=\mathbf{F}_{g-1}$.
The mapping $\mathcal{H}_g$ in
Section~\ref{subsec:cross_domain_rg}
consists of this block sequence followed by a complex
convolution applied to the last block's output.}

Following the channel-attention mechanism in
RCAN~\cite{zhang2018image}, a channel descriptor is obtained
by applying adaptive average pooling separately to the real
and imaginary parts of the complex feature map and
recombining them into a complex representation.
\added{The resulting descriptor
$\boldsymbol{\zeta}_{g,b}\in\mathbb{C}^{C\times1\times1}$
is the average of $\mathbf{U}_{g,b}$ over the
$N_c\times N_t$ grid.}

\added{A bottleneck gating network maps this descriptor to
channel-wise attention weights:}
\begingroup
\ifrevisionmarked
    \color{black}
\fi
\begin{equation}
    \mathbf{s}_{g,b}
    =
    f\!\left(
        \mathbf{W}^{(u)}_{g,b}
        \!\left(
            \operatorname{CReLU}\!\left(
                \mathbf{W}^{(d)}_{g,b}
                \!\left(\boldsymbol{\zeta}_{g,b}\right)
            \right)
        \right)
    \right),
    \label{eq:crcab_attention_weights}
\end{equation}
\endgroup
\added{where $\mathbf{W}^{(d)}_{g,b}$ and
$\mathbf{W}^{(u)}_{g,b}$ denote learnable complex-valued
$1\times1$ convolutional projections, including their biases.
They reduce and restore the channel dimension, respectively.
The gating function applies sigmoid separately to the real
and imaginary components:}
\begingroup
\ifrevisionmarked
    \color{black}
\fi
\begin{equation}
    f(\cdot)
    =
    \operatorname{sigmoid}\!\left(\Re\{\cdot\}\right)
    +
    j\,\operatorname{sigmoid}\!\left(\Im\{\cdot\}\right).
    \label{eq:crcab_complex_sigmoid}
\end{equation}
\endgroup

\added{Each complex weight lies in the first quadrant and
can change both the magnitude and phase of its corresponding
residual-feature channel.}
Then the attention weights are used to
\added{reweight}
the residual feature in a channel-wise manner,
and the CRCAB output is given by
\begin{equation}
    \mathbf{F}_{g,b}
    =
    \mathbf{F}_{g,b-1}
    +
    \left(
        \mathbf{s}_{g,b}\odot\mathbf{U}_{g,b}
    \right),
    \label{eq:crcab_output}
\end{equation}
\added{where $\odot$ denotes element-wise multiplication,
with the channel weights broadcast over the
$N_c\times N_t$ grid.}

\subsubsection{PRCAB}

\begin{figure}[!t]
    \centering
    \includegraphics[width=0.5\textwidth]{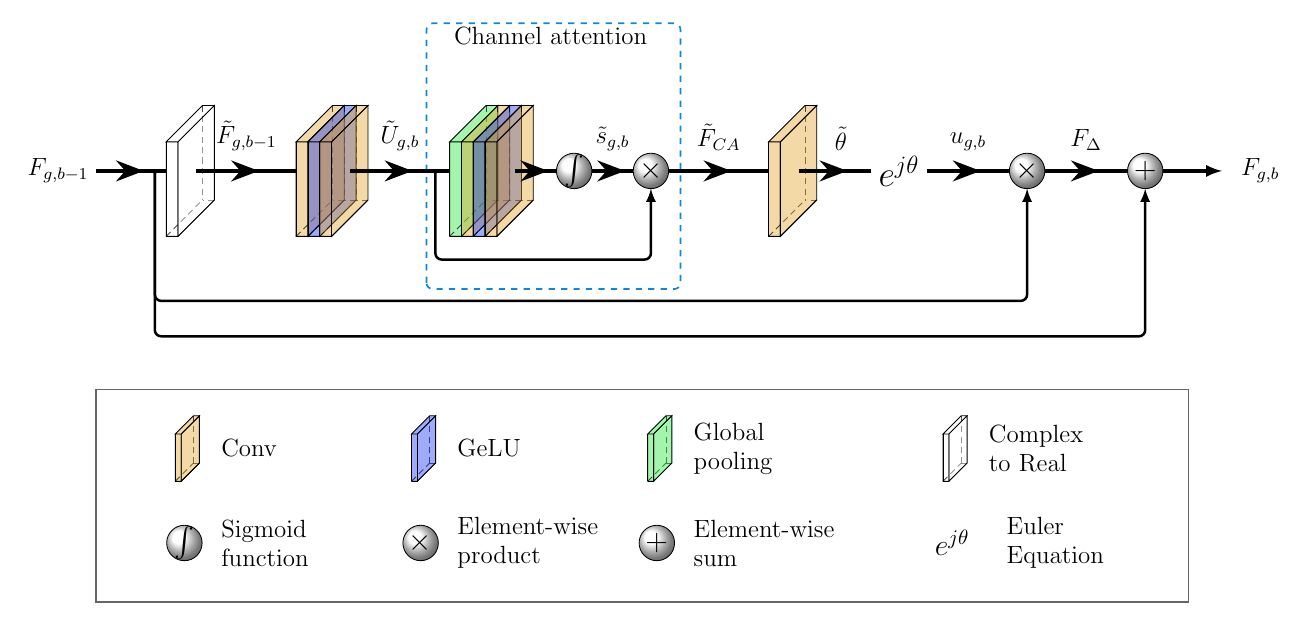}
    \caption{Phase-Rotation Residual Channel Attention Block
    (PRCAB).}
    \label{fig:PRCAB}
\end{figure}

\added{The sample-wise multiplication by PN in
\eqref{eq:joint_time_domain_model} motivates the
input-dependent multiplicative feature update in PRCAB,
complementing the additive residual refinement in CRCAB.}
The overall architecture of PRCAB is illustrated in
Fig.~\ref{fig:PRCAB}.

\added{PRCAB uses a real-valued auxiliary branch for angle
regression while retaining a complex-valued main feature
stream.}
Specifically, the complex-valued input feature
$\mathbf{F}_{g,b-1}\in\mathbb{C}^{C\times N_c\times N_t}$
is first mapped to a real-valued representation by
concatenating its real and imaginary parts along the channel
dimension, yielding
$\tilde{\mathbf{F}}_{g,b-1}
\in\mathbb{R}^{2C\times N_c\times N_t}$.

\added{Two real-valued $1\times1$ convolutions with an
intermediate \textcolor{RevisionCurrent}{Gaussian error linear unit (GELU) activation~\cite{hendrycks2016gaussian}} produce the auxiliary feature
$\tilde{\mathbf{U}}_{g,b}
\in\mathbb{R}^{C\times N_c\times N_t}$.
A real-valued channel-attention module then applies global
average pooling, a bottleneck with GELU activation, and
sigmoid gating to obtain
$\tilde{\mathbf{s}}_{g,b}\in(0,1)^{C\times1\times1}$.
The channel-attended feature is}
\begin{equation}
    \tilde{\mathbf{F}}_{\mathrm{CA}}
    =
    \tilde{\mathbf{s}}_{g,b}
    \odot
    \tilde{\mathbf{U}}_{g,b},
    \label{eq:prcab_real_attention}
\end{equation}
\added{where the attention weights are broadcast over the
grid. A real-valued $1\times1$ convolutional \textcolor{RevisionCurrent}{projection} maps
$\tilde{\mathbf{F}}_{\mathrm{CA}}
\in\mathbb{R}^{C\times N_c\times N_t}$
to an angle map:}
\begingroup
\ifrevisionmarked
    \color{black}
\fi
\begin{equation}
    \tilde{\boldsymbol{\theta}}_{g,b}
    =
    \theta_{\max}
    \tanh\!\left(
        \mathcal{W}_{\theta}\!\left(
            \tilde{\mathbf{F}}_{\mathrm{CA}}
        \right)
    \right),
    \label{eq:prcab_angle}
\end{equation}
\endgroup
\added{where $\mathcal{W}_{\theta}$ is the learnable
projection and $\theta_{\max}$ is a fixed angle bound.
The map
$\tilde{\boldsymbol{\theta}}_{g,b}
\in\mathbb{R}^{C\times N_c\times N_t}$
assigns an angle to each channel and grid position.
These angles parameterize feature updates rather than
direct estimates of the oscillator phase.}

\added{The predicted angles define unit-modulus phasors,
which act element-wise on the input features.
The corresponding residual is}
\begingroup
\ifrevisionmarked
    \color{black}
\fi
\begin{equation}
    \mathbf{F}_{\Delta}
    =
    \mathbf{F}_{g,b-1}
    \odot
    \exp\!\left(
        j\tilde{\boldsymbol{\theta}}_{g,b}
    \right)
    -
    \mathbf{F}_{g,b-1},
    \label{eq:prcab_residual}
\end{equation}
\endgroup
and the block output is given by
\begingroup
\ifrevisionmarked
    \color{black}
\fi
\begin{equation}
    \begin{aligned}
        \mathbf{F}_{g,b}
        &=
        \mathbf{F}_{g,b-1}
        +
        \boldsymbol{\eta}_{g,b}\odot\mathbf{F}_{\Delta}
        \\
        &=
        \mathbf{F}_{g,b-1}\odot
        \Bigl[
            1+\boldsymbol{\eta}_{g,b}\odot
            \bigl(
                e^{j\tilde{\boldsymbol{\theta}}_{g,b}}-1
            \bigr)
        \Bigr],
    \end{aligned}
    \label{eq:prcab_output}
\end{equation}
\endgroup
\added{where
$\boldsymbol{\eta}_{g,b}
\in(-\eta_{\max},\eta_{\max})^{C\times1\times1}$
is obtained by applying $\eta_{\max}\tanh(\cdot)$ to
learnable real-valued channel parameters.
The gate is independent of the input and is broadcast over
the grid, controlling the strength and sign of the update.}

\added{Although the predicted phasor has unit magnitude,
the complete gated factor generally does not.
Consequently, PRCAB can modify both feature magnitude and
phase.
The angle bound and the gate bound jointly limit the
rotation applied by each block; their values are specified
in \textcolor{RevisionCurrent}{Section~\ref{sec:experimental_setup}}.}

\par
\endgroup

\begingroup
\ifrevisionmarked
    \setaddedmarkup{\textcolor{black}{#1}}
\fi

\subsection{Channel and PN Estimation Tails}
\label{subsec:dual_output_heads}

\begingroup
\color{RevisionCurrent}

The channel tail maps the final frequency-domain features
$\mathbf{F}_G$ to $\widehat{\mathbf{H}}$ through $\operatorname{CConv}$ and $\operatorname{CReLU}$, reducing the feature
channels to one while preserving the $N_c\times N_t$ grid.
The PN tail concatenates $\mathbf{F}_{G-1}$ with
$\mathcal{IFFT}(\mathbf{F}_G)$ and applies two $\operatorname{CConv}$ to obtain
$\mathbf{V}\in\mathbb{C}^{C_V\times N_c\times N_t}$.
Both inputs are thus represented along the time-domain
sample axis.
These features $\mathbf{V}$ are combined with the received
grid $\mathbf{Y}$, known pilot grid $\mathbf{X}_{\mathrm{pil}}$,
and channel estimate $\widehat{\mathbf{H}}$ to refine PN.

Low-order Fourier representations have been used for
PN compensation~\cite{petrovic2007effects}, while
pilot-assisted LS methods estimate compensation parameters
from known symbols~\cite{zou2007compensation,leshem2017phase}.
We represent the sample-wise inverse
$q_m[n]=1/p_{\mathrm{eff},m}[n]$, wherever the denominator
is nonzero, using the Fourier modes
$\mathcal{I}_q=\{-R_q,\ldots,R_q\}$:
\begin{equation}
    q_m[n]
    \approx
    \frac{1}{\sqrt{N_c}}
    \sum_{d\in\mathcal{I}_q}
    c_m[d]e^{j2\pi nd/N_c},
    \label{eq:inverse_pn_basis}
\end{equation}
where $c_m[d]\in\mathbb{C}$ is the coefficient of Fourier
mode $d$ in the inverse-PN representation of OFDM symbol $m$,
and $K_q=2R_q+1\leq N_c$ is the number of retained coefficients.
In vector form, $\mathbf{q}_m\approx\mathbf{B}\mathbf{c}_m$,
where $\mathbf{B}\in\mathbb{C}^{N_c\times K_q}$ contains
the corresponding unitary Fourier basis vectors and
$\mathbf{c}_m\in\mathbb{C}^{K_q}$ collects these coefficients.
Hence,
$\mathbf{B}^{\mathrm H}\mathbf{B}=\mathbf{I}_{K_q}$,
where ${}^{\mathrm H}$ denotes conjugate transpose.
Multiplying the received time-domain samples by $q_m[n]$
corresponds to circular convolution of $Y_m[k]$ with
$c_m[d]/\sqrt{N_c}$ across subcarriers.
These scaled coefficients therefore form a
$K_q$-tap frequency-domain compensation filter.
Inverse PN compensation targets the channel-weighted
symbols $H_m[k]X_m[k]$ in
\eqref{eq:joint_frequency_domain_model}, following the
multiplicative relation in \eqref{eq:joint_time_domain_model}.
At the known pilot positions $\mathcal{K}_m$, the compensated
observations are fitted to
$\widehat{H}_m[k][\mathbf{X}_{\mathrm{pil}}]_{k,m}$,
giving the linear system
\begin{equation}
    \begin{aligned}
        \mathbf{A}_m\mathbf{c}_m
        &\approx \mathbf{b}_m,\\
        [\mathbf{A}_m]_{k,d}
        &= \frac{Y_m[(k-d)_{N_c}]}{\sqrt{N_c}},\\
        [\mathbf{b}_m]_k
        &= \widehat{H}_m[k]
           [\mathbf{X}_{\mathrm{pil}}]_{k,m},
        \quad k\in\mathcal{K}_m,\ d\in\mathcal{I}_q .
    \end{aligned}
    \label{eq:pilot_inverse_pn_system}
\end{equation}
Here, the rows of
$\mathbf{A}_m\in\mathbb{C}^{|\mathcal{K}_m|\times K_q}$
are indexed by the pilot positions and its columns by the
retained Fourier modes.

We use a regularized LS estimate centered on the
unit-compensation coefficients $\mathbf{c}_{\mathrm{uni}}$:
\begin{equation}
    \mathbf{c}_m^{\mathrm{LS}}(\lambda)
    =
    \underset{\mathbf{c}}{\operatorname{arg\,min}}
    \left\{
        \|\mathbf{A}_m\mathbf{c}-\mathbf{b}_m\|_2^2
        +
        \lambda\nu_m
        \|\mathbf{c}-\mathbf{c}_{\mathrm{uni}}\|_2^2
    \right\},
    \label{eq:pn_regularized_ls_objective}
\end{equation}
where $\lambda>0$ and
$\nu_m=\|\mathbf{A}_m\|_F^2/K_q$ scales the regularization
relative to the observation-matrix energy.
The vector $\mathbf{c}_{\mathrm{uni}}$ has a DC coefficient
of $\sqrt{N_c}$ and zero coefficients elsewhere, so that
$\mathbf{B}\mathbf{c}_{\mathrm{uni}}=\mathbf{1}$.
It therefore represents leaving the received samples
unchanged.
The first term fits the compensated pilot observations,
while the second limits departures from this reference.
For $\nu_m>0$, the solution is
\begin{equation}
    \begin{aligned}
        \mathbf{c}_m^{\mathrm{LS}}(\lambda)
        ={}&
        \mathbf{c}_{\mathrm{uni}}
        +
        \left(
            \mathbf{A}_m^{\mathrm H}\mathbf{A}_m
            +\lambda\nu_m\mathbf{I}_{K_q}
        \right)^{-1}
        \\
        &\quad{}\times
        \mathbf{A}_m^{\mathrm H}
        \left(
            \mathbf{b}_m
            -\mathbf{A}_m\mathbf{c}_{\mathrm{uni}}
        \right).
    \end{aligned}
    \label{eq:pn_regularized_ls}
\end{equation}
If $\nu_m=0$, the estimate is set to
$\mathbf{c}_{\mathrm{uni}}$.

\textcolor{RevisionCurrent}{Following model-driven learning and algorithm unrolling~\cite{he2019model,monga2021algorithm}, the Unfolded PN Refinement Tail combines relaxed LS updates with learned corrections over a fixed number of stages.}
Let
$\mathbf{C}_q^{\mathrm{LS}}(\lambda)
=[\mathbf{c}_0^{\mathrm{LS}}(\lambda),\ldots,
\mathbf{c}_{N_t-1}^{\mathrm{LS}}(\lambda)]
\in\mathbb{C}^{K_q\times N_t}$.
Starting from
$\mathbf{C}_q^{(0)}
=\mathbf{C}_q^{\mathrm{LS}}(\lambda_0)$,
stage $i=1,\ldots,J$ computes
\begin{equation}
    \begin{aligned}
        \overline{\mathbf{C}}_q^{(i)}
        ={}&
        (1-\alpha_i)\mathbf{C}_q^{(i-1)}
        +\alpha_i\mathbf{C}_q^{\mathrm{LS}}(\lambda_i),
        \\
        \mathbf{C}_q^{(i)}
        ={}&
        \overline{\mathbf{C}}_q^{(i)}
        +
        \mathbf{B}^{\mathrm H}
        \mathcal{N}_i
        \left(
            \left[
                \mathbf{B}\overline{\mathbf{C}}_q^{(i)},
                \mathbf{V}
            \right]
        \right).
    \end{aligned}
    \label{eq:pn_unfolded_updates}
\end{equation}
The initialization parameter $\lambda_0$ is fixed, whereas
$\alpha_i\in(0,1)$ and $\lambda_i>0$ are learnable scalars.
The first update moves the coefficients toward the
regularized LS solution, reintroducing pilot consistency
after the preceding learned correction.
The second update uses a stage-specific complex-valued CNN
$\mathcal{N}_i$, consisting of two convolutions separated
by $\operatorname{CReLU}$.
Its input concatenates the reconstructed inverse-PN grid
with $\mathbf{V}$ along the feature-channel dimension.
The CNN outputs one complex-valued grid, which
$\mathbf{B}^{\mathrm H}$ projects onto the retained
coefficient subspace.
The pilot systems and $\widehat{\mathbf{H}}$ remain fixed
across the stages within each forward pass, and no
data-decision feedback is used.

After the final stage,
$\widehat{\mathbf{q}}=\mathbf{B}\mathbf{C}_q^{(J)}
\in\mathbb{C}^{N_c\times N_t}$
is inverted element-wise to obtain $\mathbf{Z}_{\mathrm{PN}}$.
Entries with
$|\widehat{\mathbf{q}}|\leq\epsilon_{\mathrm{tail}}$
are assigned a reciprocal value of one, where
$\epsilon_{\mathrm{tail}}>0$ is a small fixed numerical threshold
that prevents division by near-zero values.
Amplitude-constrained formulations have also been studied
for pilot-based PN compensation
\cite{mathecken2017constrained}.
Here, a phase-preserving soft mapping controls the magnitude
of the inverse-derived PN estimate:
\begin{equation}
    \widehat{\mathbf{p}}_{\mathrm{eff}}
    =
    \frac{\mathbf{Z}_{\mathrm{PN}}}
         {\mathbf{R}_{\epsilon_{\mathrm{tail}}}}
    \odot
    \left[
        1+\Delta_\rho
        \tanh\!\left(
            \frac{
                \gamma_\rho
                (\mathbf{R}_{\epsilon_{\mathrm{tail}}}-1)
            }{\Delta_\rho}
        \right)
    \right],
    \label{eq:pn_soft_normalization}
\end{equation}
where
$\mathbf{R}_{\epsilon_{\mathrm{tail}}}
=\max(|\mathbf{Z}_{\mathrm{PN}}|,\epsilon_{\mathrm{tail}})$
uses the same threshold to safeguard magnitude normalization,
and all operations are element-wise.
The global learnable scalars
$0<\Delta_\rho<1$ and $0<\gamma_\rho<1$
control the magnitude range and the slope at unit magnitude,
respectively.
For entries with
$|\mathbf{Z}_{\mathrm{PN}}|\geq\epsilon_{\mathrm{tail}}$,
the mapping preserves phase and contracts magnitude
deviations from one, with output magnitude in
$(1-\Delta_\rho,1+\Delta_\rho)$.
The PN target in \eqref{eq:effective_pn_target} need not
have unit magnitude after receive filtering, as illustrated
by \eqref{eq:pn_weighted_phasor}.
This motivates allowing bounded departures from unit
modulus in the output mapping.
The supervision targets themselves are not clipped.
The PN loss also propagates through the LS computations
to the channel tail and shared backbone.
The stage count, coefficient order, CNN configuration,
and parameter settings are given in
Section~\ref{sec:experimental_setup}.

\par
\endgroup

\section{Experimental Setup}
\label{sec:experimental_setup}

This section describes the simulation configuration,
data generation and supervision targets, network configuration
and training methodology, benchmark schemes, and evaluation
protocol.
The simulations follow the waveform-level SISO CP-OFDM system
in Section~II, including pulse shaping, time-varying multipath
propagation, receiver PN, matched filtering, and additive noise.
Unless otherwise specified, the following main configuration
is used; parameter variations are stated in the corresponding
experiments.

\subsection{Simulation Configuration}
\label{subsec:sim_setup}

The system operates at a carrier frequency of $100$~GHz with
$N_c=2048$ subcarriers and a subcarrier spacing of
$\Delta f=480$~kHz, giving an OFDM frequency-grid bandwidth of
$983.04$~MHz.
Each observation block contains $N_t=8$ consecutive OFDM symbols.
Uncoded Gray-mapped 16-QAM is used for data, whereas QPSK is
used for pilots.
Both constellations are normalized to unit average symbol energy.
A CP of $512$ samples is used in both propagation scenarios,
corresponding to $0.52083~\mu\mathrm{s}$ and $20\%$ of the total
OFDM symbol duration.
The main waveform parameters are summarized in
Table~\ref{tab:sim_params}.

\begin{table}[t]
    \caption{\revnow{Main Simulation and Network Parameters}}
    \label{tab:sim_params}
    \centering
    \small
    \renewcommand{\arraystretch}{1.12}
    \setlength{\tabcolsep}{4pt}
    \begin{tabular}{>{\raggedright\arraybackslash}p{0.53\linewidth} >{\raggedright\arraybackslash}p{0.39\linewidth}}
        \hline
        Parameter & Setting \\
        \hline
        Link configuration
        & SISO CP-OFDM \\

        Carrier frequency $f_c$
        & $100$~GHz \\

        Number of subcarriers $N_c$
        & $2048$ \\

        Subcarrier spacing $\Delta f$
        & $480$~kHz \\

        Baseband sampling rate $N_c\Delta f$
        & $983.04$~MSa/s \\

        Useful symbol duration
        & $2.08333~\mu\mathrm{s}$ \\

        CP length $L_{\mathrm{CP}}$
        & $512$ samples \\

        CP duration
        & $0.52083~\mu\mathrm{s}$ \\

        CP fraction of total symbol duration
        & $20\%$ \\

        Total symbol duration $T_{\mathrm{sym}}$
        & $2.60417~\mu\mathrm{s}$ \\

        Observed symbols $N_t$
        & $8$ \\

        Duration of eight complete symbols
        & $20.8333~\mu\mathrm{s}$ \\

        Data/pilot modulation
        & Gray 16-QAM/QPSK \\

        Waveform oversampling factor
        & $8$ \\

        SRRC roll-off factor
        & $0.25$ \\

        Transmit and receive SRRC filters
        & $129$ coefficients each\\
        
        \revnow{RAN4 reference carrier $f_{\mathrm{base}}$}
        & \revnow{$29.55$~GHz} \\
        
        \revnow{FoM coefficient $\kappa_{\mathrm{FoM}}$}
        & \revnow{$9$~dB/decade} \\
        \hline
    \end{tabular}
\end{table}

The rotating-comb pattern in Fig.~\ref{fig:pilot_design}
uses a pilot spacing of $S_f=8$ subcarriers, with the
offset advancing by one subcarrier per OFDM symbol.
This shift gives $S_t=S_f=8$.
Each symbol contains $256$ pilots, and each subcarrier
carries a pilot once over the eight-symbol block.
The pilot values and locations are known to the receiver.
For the main channel configuration, the frequency-correlation
magnitude computed from the nominal power delay profile (PDP)
at the resulting
$3.84$-MHz pilot separation has a median of $0.943$
and a fifth percentile of $0.691$.
The first and last symbol centers are separated by
$18.23~\mu$s, over which the Jakes diffuse-fading
correlation is approximately $0.975$ at $30$~km/h
and $0.634$ at $120$~km/h.
These correlations support combining pilot information
across frequency and time, although the rotating
pattern samples different subcarriers at different
times and temporal correlation weakens at higher speeds.
The benchmark schemes retain their respective pilot
configurations, as specified in their descriptions.

The transmit waveform is generated with eightfold
oversampling.
The transmitter and receiver use a matched pair of
finite-length SRRC filters, with each pulse truncated
to $[-8T_s,8T_s]$.
Fractional-delay filters implement path delays that
are not integer multiples of the simulation sampling
interval.
The receiver compensates for the fixed group delays
of the transmit and receive filters, samples on the
time grid defined in Section~\ref{sec:system_model},
removes the CP, and applies the FFT.
Adjacent symbols before and after the observation
block are included in waveform generation to preserve
the effects of channel and filter memory across
block boundaries.

The main dataset uses the scenario-conditioned 3GPP TDL
model described in Section~\ref{sec:system_model}.
Indoor hotspot (InH) and urban microcell (UMi) scenarios
each account for $50\%$ of the dataset, with TDL-A,
TDL-B, and TDL-E selected in equal proportions within
each scenario.
TDL-A/B use the scenario's NLOS statistics, whereas
TDL-E uses its LOS statistics.
For each candidate channel, the RMS delay-spread
parameter is sampled from the corresponding lognormal
distribution in TR~38.901~\cite{3gpp38901}.
For LOS channels, the overall Rician $K$-factor and
logarithmic delay spread are sampled jointly using
the statistics in Table~7.5-6 of that report.
These sampled parameters determine the tap-power
composition and delay scaling, with the mean path
powers normalized to sum to one.
Waveform propagation uses the instantaneous path
gains, including their Doppler-induced time variation.
A candidate is accepted only if its RMS delay-spread
parameter does not exceed $100$~ns and its effective
channel memory fits within the CP.
This memory accounts for multipath delays,
fractional-delay filtering, and transmit and receive
pulse filtering.
Rejected candidates are resampled within the same
scenario and TDL profile, while accepted channels
retain all their taps.
Conditions beyond these limits are evaluated
separately in the stress tests.

The velocity distribution depends on the scenario.
InH uses static and pedestrian cases with
probabilities of $62.5\%$ and $37.5\%$.
UMi uses static, pedestrian, urban-vehicle, and
highway-vehicle cases with probabilities of
$50\%$, $30\%$, $17\%$, and $3\%$.
Static cases have zero velocity.
Pedestrian speeds in both scenarios are drawn from
a Gaussian distribution with mean $3$~km/h and
standard deviation $1$~km/h, truncated to
$[0.5,7]$~km/h.
Urban-vehicle speeds have Gaussian parameters
of $30$~km/h and $10$~km/h, respectively, and
are truncated to $[8,60]$~km/h.
Highway-vehicle speeds have Gaussian parameters
of $100$~km/h and $15$~km/h, respectively, and
are truncated to $[60,120]$~km/h.
The sampled speed determines the maximum Doppler
shift, $f_d=vf_c/c$.
The diffuse fading of each tap is generated using
the generalized method of exact Doppler spread
(GMEDS)~\cite{paetzold2006two} with $48$ sinusoidal
components.
The LOS specular component uses a Doppler shift
of $0.7f_d$ and zero initial phase at the channel
generator's time origin.

The main PN configuration uses the RAN4 parametric
spectrum described in Section~\ref{subsec:pn_model}.
Its target spectrum at $100$~GHz is obtained using
the carrier-frequency scaling and FoM rules in that
subsection, with a spectral-level adjustment of
$\Delta_{\mathrm{cal}}=-10$~dB informed by the measured
synthesizer PN reported in~\cite{liu2020synthesizer}.
The RAN1 Set1 and Set2 spectra use the same general
pole--zero formulation with different parameter sets
and carrier-frequency scaling rules.
They are used as benchmark test PN models with the same
$-10$-dB spectral-level adjustment.

The generalization tests additionally use PSDs
digitized from oscillator measurements at
$79.2$, $100.8$, and $140$~GHz
in~\cite{plouchart2013synthesizer,
liu2020synthesizer,callender2022transmitter},
respectively.
Within each measured range, the PSD values in dB
are interpolated linearly against logarithmic
frequency offset, without an additional
spectral-level adjustment.
Outside the frequency-offset ranges reported in the
original papers, each PSD is extended using  3GPP
spectral templates selected to best match its measured
spectral shape, with continuous matching at the
measured boundaries.
For the $79.2$-GHz PSD, the measured curve extends down
to $1$~kHz and
its high-frequency extension uses RAN1 Set1.
For the $100.8$-GHz PSD, RAN1 Set2 extends the curve
from the lowest measured offset of $10$~kHz down
to $1$~kHz and beyond the highest measured offset.
The $140$-GHz PSD uses the same frequency
extension procedure with RAN1 Set2.
For all PN models, the PSD below $1$~kHz is held
constant at its $1$-kHz value.
Continuous phase sequences are synthesized across
OFDM symbols using multirate finite impulse
response (FIR) spectral shaping.
The three parametric spectra and three
measurement-based spectra are shown in
Fig.~\ref{fig:pn_psd}.

\begin{figure}[!t]
    \centering
    \includegraphics[width=\linewidth]{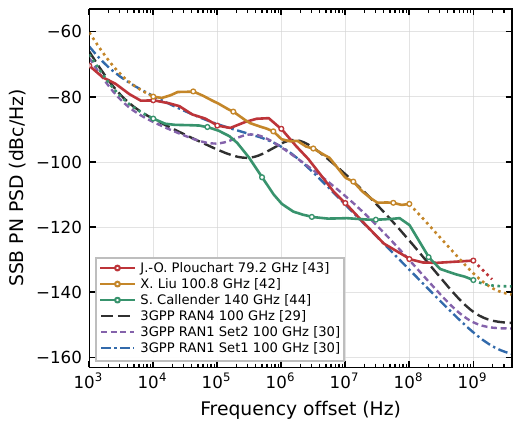}
    \caption{\textcolor{RevisionCurrent}{Prescribed SSB PN PSDs used in the experiments. The 3GPP models use $f_c=100$~GHz and $\Delta_{\mathrm{cal}}=-10$~dB; the measured spectra~\cite{plouchart2013synthesizer,liu2020synthesizer,callender2022transmitter} retain their native carriers and levels. Solid curves with circles show digitized measurement bands; dotted segments are model continuations.}}
    \label{fig:pn_psd}
\end{figure}

Complex AWGN is added before receive filtering.
The noise level is specified by the uncoded information-bit
ratio $E_b/N_0$, including pilot and CP overhead.
Let $\rho_{\mathrm{pil}}$ denote the pilot resource fraction.
The target received-grid signal-to-noise ratio (SNR) is set to
\begin{equation}
    \gamma_{\mathrm{grid}}
    =
    \frac{E_b}{N_0}
    \log_2(M_{\mathrm{QAM}})
    (1-\rho_{\mathrm{pil}})
    \frac{N_c}{N_c+L_{\mathrm{CP}}},
    \label{eq:setup_ebn0_grid_snr}
\end{equation}
where $M_{\mathrm{QAM}}$ denotes the data-constellation
size, with $M_{\mathrm{QAM}}=16$ for training and
$\rho_{\mathrm{pil}}=1/8$ in the main configuration.
Numerical noise calibration uses the average power of each
noiseless received FFT grid after channel propagation, PN,
and receive filtering.
The prefilter noise variance is obtained by accounting for
the receive-filter energy and the FFT scaling convention.
The training and evaluation ranges of $E_b/N_0$ are specified
in the corresponding subsections.

\subsection{Dataset Generation and Supervision Targets}
\label{subsec:dataset_generation}

Each frame comprises $N_t=8$ observed OFDM symbols
extracted from a continuous transmitted waveform.
Channels are generated according to the distributions and
acceptance criteria in Section~\ref{subsec:sim_setup}.
Separate random-number streams control channel fading,
transmitted data, PN, and additive noise.
\textcolor{RevisionCurrent}{The parameters and random seeds associated with each sample are retained so that its received observations and supervision targets can be reproduced.}

Training uses sample banks of 12,800 frames, with
$E_b/N_0$ sampled uniformly over $20$--$30$~dB.
Each bank is shuffled and traversed once per epoch
and replaced every 20 epochs.
All frame realizations, including noise and the
assigned $E_b/N_0$, remain fixed between replacements.
Test sets are generated using random seeds distinct
from those used for training.
Unless explicitly varied, their settings follow
Section~\ref{subsec:sim_setup}.
Experiment-specific channel and PN models, parameter
settings, and test-set sizes are specified alongside
the corresponding results.

Supervision uses the reference channel $\mathbf{H}$
and PN $\mathbf{p}_{\mathrm{eff}}$ defined in
Section~\ref{subsec:problem_formulation}, both
constructed without additive noise.
\revnow{%
When $|\bar y_m[n]|\leq10^{-12}$, the PN target is set to
$1$; otherwise, the ratio in \eqref{eq:effective_pn_target}
is used.%
}
The PN targets are neither projected onto the unit
circle nor magnitude-clipped.
Checkpoint selection uses a fixed validation set
spanning $E_b/N_0=0$--$30$~dB.
The checkpoint with the lowest normalized area under
the validation $\log_{10}\mathrm{BER}$ versus
$E_b/N_0$ curve is selected for testing, where BER
denotes the bit error rate.

\subsection{Network Configuration and Training Methodology}
\label{subsec:training_methodology}

The main model uses four alternating IFFT/FFT residual
groups, each containing five blocks, with 64 feature
channels and $3\times3$ spatial convolution kernels.
\textcolor{RevisionCurrent}{%
The third block in each residual group is a PRCAB;
all remaining blocks are CRCABs.
The PRCAB bounds are $\theta_{\max}=\pi/2$ and
$\eta_{\max}=0.2$.%
}
\textcolor{RevisionCurrent}{%
The Unfolded PN Refinement Tail uses $J=3$ stages,
$K_q=65$ Fourier coefficients, and $C_V=32$ auxiliary
feature channels.
Each stage contains a complex-valued CNN with 32 hidden
channels and $3\times3$ kernels.
The initial LS regularization is fixed at $\lambda_0=0.1$.
The reciprocal and magnitude-normalization safeguards use
$\epsilon_{\mathrm{tail}}=10^{-8}$.
The soft-mapping parameters are learned jointly with
the network.%
}
Variations of these architectural choices are examined
in the ablation study.

Let $\mathcal{L}_H$ and $\mathcal{L}_p$ denote the squared
complex estimation errors for the reference channel and
\textcolor{RevisionCurrent}{PN}, respectively, averaged
over the grid and minibatch.
The uncertainty-weighted joint loss~\cite{kendall2018multi}
is
\begin{equation}
    \mathcal{L}_{\mathrm{joint}}
    =
    e^{-s_H}\mathcal{L}_H
    +
    e^{-s_p}\mathcal{L}_p
    +
    \frac{1}{2}(s_H+s_p).
    \label{eq:training_joint_loss}
\end{equation}
The global learnable task-weight parameters are initialized
as $s_H=s_p=0$.
Their learning rate follows that of the network parameters.

Training uses AdamW~\cite{loshchilov2017decoupled} with an
effective minibatch size of 32 and network weight decay
of $10^{-4}$; no weight decay is applied to $s_H$ or $s_p$.
\textcolor{RevisionCurrent}{Training lasts 100 epochs.}
A one-epoch warm-up~\cite{goyal2017accurate} raises the
learning rate from $10^{-6}$ to $10^{-4}$, followed by cosine
decay~\cite{loshchilov2016sgdr}.
\textcolor{RevisionCurrent}{%
After 60 epochs, a second one-epoch warm-up raises the
learning rate from its current value to $2\times10^{-4}$,
followed by cosine decay to $10^{-5}$ over the remaining
39 epochs.%
}
The optimizer state is retained throughout.
The checkpoint with the lowest normalized area under the
validation $\log_{10}\mathrm{BER}$ versus $E_b/N_0$ curve
is selected for testing.

\subsection{Benchmark Schemes}
\label{subsec:benchmark_schemes}

\revnow{%
The receivers are evaluated using paired channel and PN
realizations, with their respective pilot and data allocations.
Pilot overhead is measured over the useful OFDM grid,
excluding CP; noise calibration includes each configuration's
actual pilot and CP costs.
The iterative LS and cascaded receivers use a full-pilot
first symbol followed by fixed comb pilots with spacing eight.
Their overhead is $23.44\%$ for eight symbols, lies in
$(12.5\%,23.44\%]$ for frames of at least eight symbols,
and approaches $12.5\%$ asymptotically.%
}

\subsubsection{Iterative LS Receiver}

\revnow{%
Following Zou et al.~\cite{zou2007compensation},
this receiver jointly estimates channel and PN from the
full-pilot symbol, then fixes the channel estimate while
alternating between data and PN estimation.
Independent calibration selects channel lags $[-16,495]$
and a piecewise-linear complex-PN model with five
uniformly spaced nodes.
Each stage performs 5--50 iterations, stopping when
relative objective improvement does not exceed $10^{-5}$
for three consecutive iterations.
Data updates use continuous-valued LS without hard
decisions or decoder feedback.%
}

\subsubsection{DeepCascade Receiver}

\revnow{%
The cascaded receiver of
Mohammadian et al.~\cite{mohammadian2021deep}
uses ChDNN, PnDNN, and DnCNN to refine the channel,
inverse-PN coefficients, and data symbols.
For FFT2048, ChDNN learns a residual correction to
the full channel impulse response (CIR) estimate,
PnDNN refines five complex
coefficients, and DnCNN denoises compensated symbols
reshaped to $32\times64$.
Channel estimation is frame-level and PN compensation
is symbol-wise.
Training uses CRFCAN's channel/PN distributions and
20--30~dB range; the budget, optimizer, minibatch size,
and learning-rate schedule follow
Section~\ref{subsec:training_methodology}.
Stage-specific MSE losses are equally weighted,
with gradients detached between stages.
PN errors are evaluated over the seven data-bearing symbols.%
}

\subsubsection{CP-OFDM Adaptation of DeepSRX}

\revnow{%
DeepSRX~\cite{marasinghe2025phase},
originally a single-carrier neural transceiver,
is adapted by applying its residual convolutions along
OFDM subcarriers with fixed 16QAM and no transmitter-side
constellation or pilot optimization.
It uses CRFCAN's rotating-comb pilots with $12.5\%$ overhead.
Inputs comprise the real and imaginary parts of received
symbols and interpolated LS channel estimates,
together with the pilot mask.
Symbol-wise soft-bit outputs are trained using
binary cross-entropy.
The original Adam optimizer, fixed $10^{-3}$ learning rate,
and 8--30~dB training range are retained;
training lasts 100 epochs with minibatches of ten frames.
As it does not explicitly estimate channel or PN,
DeepSRX is evaluated only through BER.%
}

\subsubsection{Distributed-PTRS Receiver (PTRS-CPE)}

\revnow{%
Following Qi et al.~\cite{qi2018ptrs},
distributed phase-tracking reference signals (PTRS)
are inserted every 24 subcarriers alongside
rotating-comb channel pilots, giving $16.17\%$
total overhead after accounting for overlaps.
Phase-aligned comb-LS estimates are stitched into
a frame-level channel estimate, and channel-weighted
PTRS correlation estimates symbol-wise CPE.
CPE correction precedes channel equalization,
without PN-amplitude correction or explicit ICI suppression.
The PTRS pattern and phase-tracking procedure are adapted
to the adopted SISO CP-OFDM grid.%
}

\subsubsection{Block-PTRS Receiver (PTRS-block)}

\revnow{%
Combining Levanen et al.'s block-PTRS
pattern~\cite{levanen2021mobile} with
Syrj\"al\"a et al.'s non-iterative PN
estimator~\cite{syrjala2019pilot},
this receiver adds 48 consecutive PTRS per symbol,
giving $14.55\%$ total overhead.
Channel reconstruction follows the distributed-PTRS receiver.
After equalization, the central 40 PTRS observations
provide an unregularized LS estimate of CPE and
four PN components on each side, followed by
conjugate PN matched-filter compensation without
gain normalization.
Adaptation concerns the grid dimensions,
reference layout, and channel-estimation interface;
the PTRS configuration is fixed before evaluation.%
}

An additional genie-aided reference uses the true channel
and PN targets for sample-wise PN compensation followed
by subcarrier-wise channel equalization.
It is evaluated on the same noisy waveform and data
positions as the corresponding receiver configuration.

\subsection{Evaluation Metrics}
\label{sec:evaluation_metrics}
Channel and PN estimation are evaluated using normalized
mean squared error (NMSE).
Let $\iota$ index the evaluated frames at each $E_b/N_0$
point within the specified test setting. The metrics are
\begin{equation}
    \begin{aligned}
        \mathrm{NMSE}_{H}
        &=
        10\log_{10}
        \frac{
            \sum_{\iota}
            \|\widehat{\mathbf H}_{\iota}-\mathbf H_{\iota}\|_F^2
        }{
            \sum_{\iota}\|\mathbf H_{\iota}\|_F^2
        },
        \\
        \mathrm{NMSE}_{\mathrm{PN}}
        &=
        10\log_{10}
        \frac{
            \sum_{\iota}
            \|\widehat{\mathbf p}_{\mathrm{eff},\iota}
              -\mathbf p_{\mathrm{eff},\iota}\|_F^2
        }{
            \sum_{\iota}\|\mathbf p_{\mathrm{eff},\iota}\|_F^2
        }.
    \end{aligned}
    \label{eq:evaluation_metrics}
\end{equation}
The two metrics are referred to as H-NMSE and PN-NMSE,
respectively, in the text and figures.
NMSE is reported in decibels without phase or scale alignment.

Data detection is evaluated using
$\mathrm{BER}=N_{\mathrm{err}}/N_{\mathrm{data}}$, where
$N_{\mathrm{err}}$ and $N_{\mathrm{data}}$ are the total
numbers of incorrectly detected and transmitted data bits,
respectively, excluding pilots.
The error vector magnitude (EVM) is
\begin{equation}
    \mathrm{EVM}
    =
    \sqrt{
        \frac{
            \sum_{\iota}
            \|\widehat{\mathbf X}_{\iota}^{\mathrm{data}}
              -\mathbf X_{\iota}^{\mathrm{data}}\|_F^2
        }{
            \sum_{\iota}\|\mathbf X_{\iota}^{\mathrm{data}}\|_F^2
        }
    }\times100\%,
    \label{eq:evm_metric}
\end{equation}
where $\mathbf X_{\iota}^{\mathrm{data}}$ and
$\widehat{\mathbf X}_{\iota}^{\mathrm{data}}$ contain the
transmitted and recovered data symbols, respectively,
with the latter taken before hard decisions.

PN-NMSE is evaluated over all eight symbols for CRFCAN,
Iterative LS, PTRS-CPE, and PTRS-block, and over the seven
data-bearing symbols for DeepCascade.
DeepSRX produces bit predictions rather than explicit
channel or PN estimates and therefore enters only the
BER comparison.
CRFCAN applies time-domain PN compensation, FFT, and
one-tap channel equalization.
The genie-aided reference follows the same procedure
using the true channel and PN targets instead of their estimates.
Iterative LS uses the data-symbol estimates from its final
LS update, while DeepCascade uses the symbols refined
by its final DnCNN stage.
PTRS-CPE corrects symbol-wise CPE before channel
equalization, whereas PTRS-block applies conjugate
PN matched-filter compensation after channel equalization.
These symbol-output receivers use nearest-neighbour
constellation demapping to obtain data-bit decisions;
DeepSRX instead thresholds its output bit logits at zero.

\section{Numerical Results and Discussion}
\label{sec:numerical_results}

This section evaluates CRFCAN in terms of joint channel
and PN estimation and data detection.
Ablation studies, tests under different propagation and
oscillator conditions, and inference-cost comparisons
are used to examine its design choices and applicability.
Unless otherwise stated, the experiments follow the
configuration in Section~\ref{sec:experimental_setup};
changes to this configuration are specified in the
corresponding subsections.

\begin{figure*}[!t]
    \centering

    \begin{subfigure}[t]{0.32\textwidth}
        \centering
        \includegraphics[width=\linewidth]
            {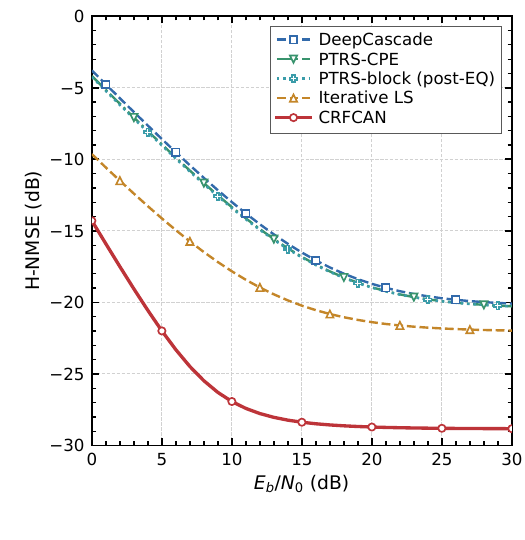}
        \caption{H-NMSE: TDL-A/B/E and 3GPP RAN4}
        \label{fig:h_nmse_matched}
    \end{subfigure}
    \hfill
    \begin{subfigure}[t]{0.32\textwidth}
        \centering
        \includegraphics[width=\linewidth]
            {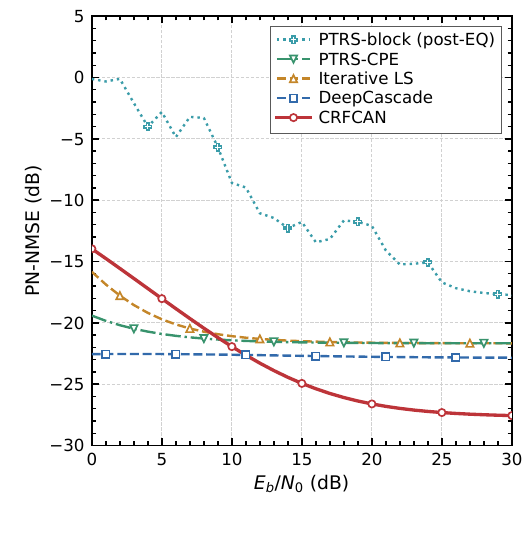}
        \caption{PN-NMSE: TDL-A/B/E and 3GPP RAN4}
        \label{fig:pn_nmse_matched}
    \end{subfigure}
    \hfill
    \begin{subfigure}[t]{0.32\textwidth}
        \centering
        \includegraphics[width=\linewidth]
            {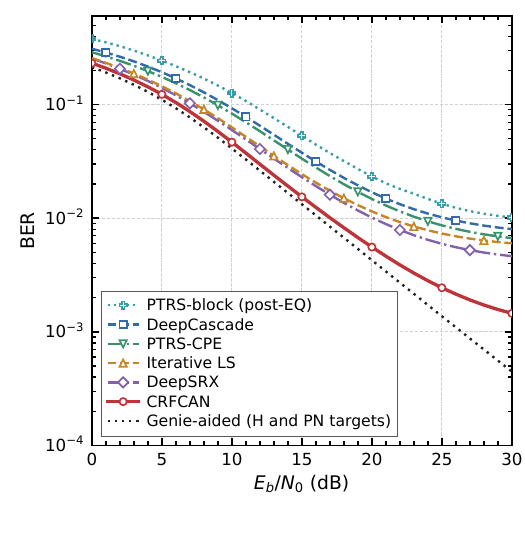}
        \caption{BER: TDL-A/B/E and 3GPP RAN4}
        \label{fig:ber_matched}
    \end{subfigure}

    \par\medskip

    \begin{subfigure}[t]{0.32\textwidth}
        \centering
        \includegraphics[width=\linewidth]
            {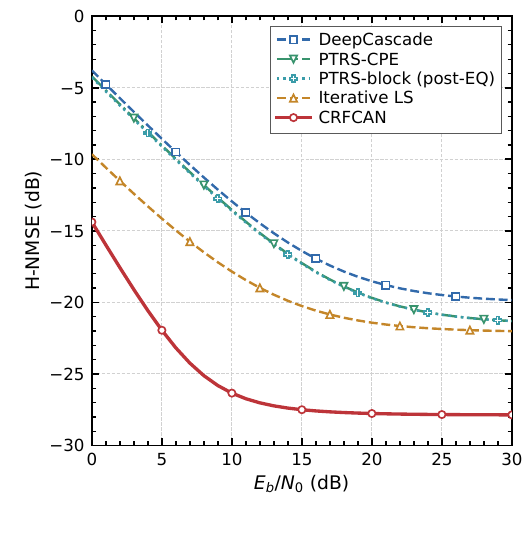}
        \caption{H-NMSE: TDL-C/D and 3GPP RAN1 Set1/2}
        \label{fig:h_nmse_generalization}
    \end{subfigure}
    \hfill
    \begin{subfigure}[t]{0.32\textwidth}
        \centering
        \includegraphics[width=\linewidth]
            {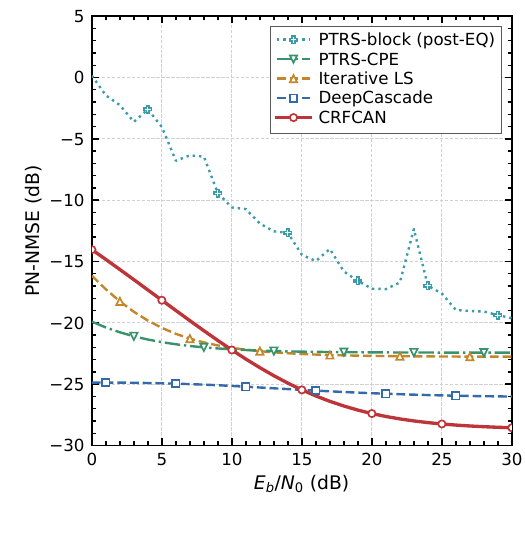}
        \caption{PN-NMSE: TDL-C/D and 3GPP RAN1 Set1/2}
        \label{fig:pn_nmse_generalization}
    \end{subfigure}
    \hfill
    \begin{subfigure}[t]{0.32\textwidth}
        \centering
        \includegraphics[width=\linewidth]
            {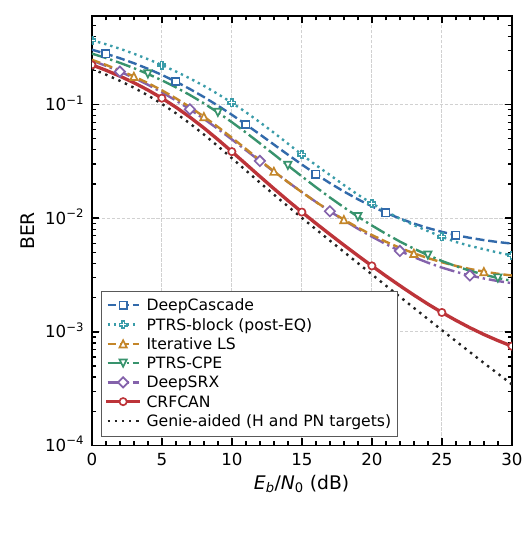}
        \caption{BER: TDL-C/D and 3GPP RAN1 Set1/2}
        \label{fig:ber_generalization}
    \end{subfigure}

    \caption{Channel and PN estimation NMSE and BER for the TDL channel and 3GPP PN. Each $E_b/N_0$ point uses 2,000 frames.}
    \label{fig:mse_3pn}
    \label{fig:ber_3pn}
\end{figure*}

\subsection{Joint Estimation and Detection Performance}
\label{subsec:joint_estimation_performance}

CRFCAN is trained over $E_b/N_0\in[20,30]$~dB and
evaluated over $E_b/N_0\in[0,30]$~dB in two test settings.
The training-matched channel/PN setting uses TDL-A/B/E
profiles in equal proportions with RAN4 PN.
The within-family generalization setting uses TDL-C/D
profiles in equal proportions, with each channel realization
evaluated under both RAN1 Set1 and Set2 PN.
The two PN models contribute equally, and their error
counts and energy sums are pooled to compute the
reported metrics.
For each learned receiver, the same trained checkpoint
is used in both settings without test-time fine-tuning.

The channel-estimation results in
Fig.~\ref{fig:mse_3pn} show that CRFCAN achieves the lowest
H-NMSE in both settings throughout the evaluated
$0$--$30$~dB range.
Its estimation error decreases rapidly as $E_b/N_0$
increases and gradually levels off beyond approximately
$15$--$20$~dB.
At $30$~dB, the H-NMSE reaches $-28.82$ and $-27.86$~dB
in the TDL-A/B/E + 3GPP RAN4 and TDL-C/D + 3GPP RAN1 Set1/2 generalization
settings, respectively.
These values improve upon Iterative LS, the strongest
channel-estimation baseline, by $6.86$ and $5.85$~dB.
At $30$~dB, CRFCAN's H-NMSE increases by $0.96$~dB
in the held-out setting, indicating that its
channel-estimation accuracy is largely retained
under the tested joint change in channel profiles and PN models.

The PN-estimation results show a different trend.
CRFCAN only achieves lower PN-NMSE than PTRS-block at low
$E_b/N_0$, while CRFCAN's PN-NMSE continues to decrease
across the medium-to-high $E_b/N_0$ range.
CRFCAN achieves the lowest PN-NMSE among the compared
estimators from $11$ and $15$~dB onward in the
training-matched and within-family generalization
settings, respectively.
At $30$~dB, its PN-NMSE reaches $-27.56$ and $-28.55$~dB.
Thus, unlike channel estimation, which levels off
earlier, PN estimation continues to benefit from
increasing $E_b/N_0$, with comparable high-SNR accuracy
in the two settings.

CRFCAN also achieves the lowest BER among the compared
receivers in both settings, with a larger advantage
at medium-to-high $E_b/N_0$.
At $30$~dB, its BER reaches $1.46\times10^{-3}$ and
$7.49\times10^{-4}$ in the training-matched and
within-family generalization settings, respectively.
These values are approximately $68.6\%$ and $72.1\%$
lower than those of DeepSRX, the strongest BER baseline
at this point.
The detection advantage therefore extends to the
held-out channel and PN configurations without fine-tuning.
CRFCAN remains close to the genie-aided reference
at low $E_b/N_0$, while a gap persists
at high $E_b/N_0$, showing that residual estimation
errors continue to limit detection as the noise decreases.

\begin{figure*}[!t]
    \centering

    \begin{subfigure}[t]{0.32\textwidth}
        \centering
        \includegraphics[width=\linewidth]
            {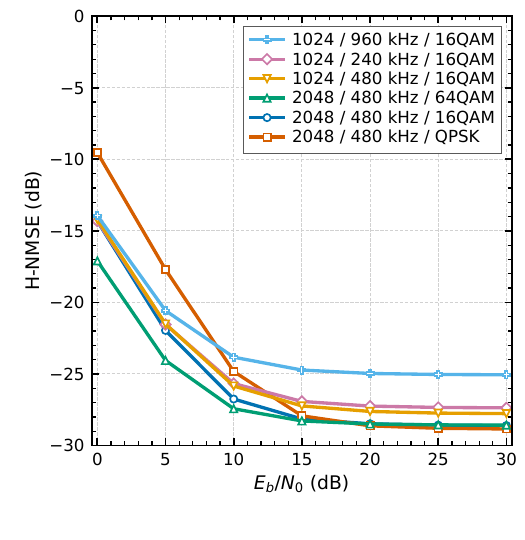}
        \caption{Modulation and numerology: H-NMSE.}
        \label{fig:gen_mod_num_h}
    \end{subfigure}
    \hfill
    \begin{subfigure}[t]{0.32\textwidth}
        \centering
        \includegraphics[width=\linewidth]
            {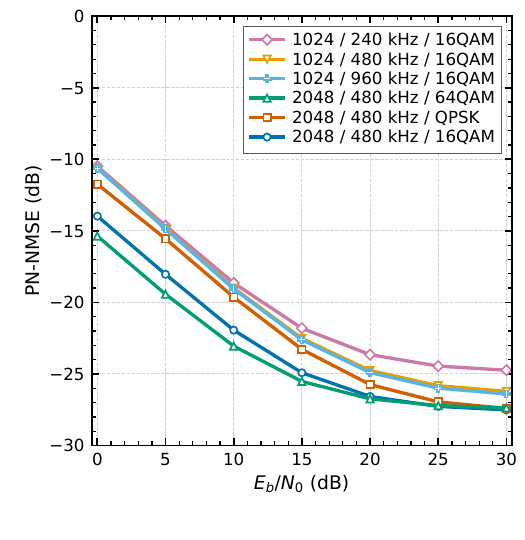}
        \caption{Modulation and numerology: PN-NMSE.}
        \label{fig:gen_mod_num_pn}
    \end{subfigure}
    \hfill
    \begin{subfigure}[t]{0.32\textwidth}
        \centering
        \includegraphics[width=\linewidth]
            {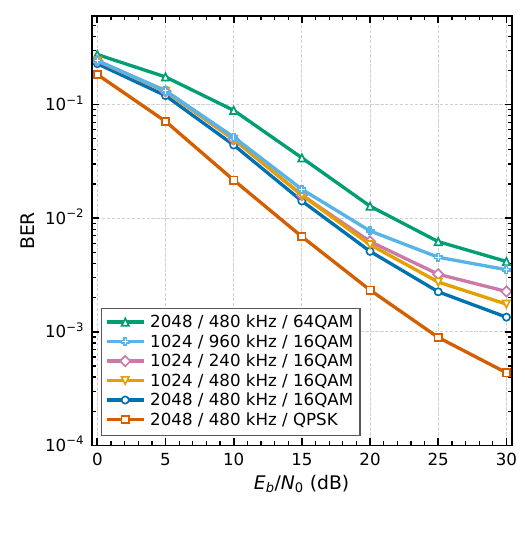}
        \caption{Modulation and numerology: BER.}
        \label{fig:gen_mod_num_ber}
    \end{subfigure}

    \par\smallskip

    \begin{subfigure}[t]{0.32\textwidth}
        \centering
        \includegraphics[width=\linewidth]
            {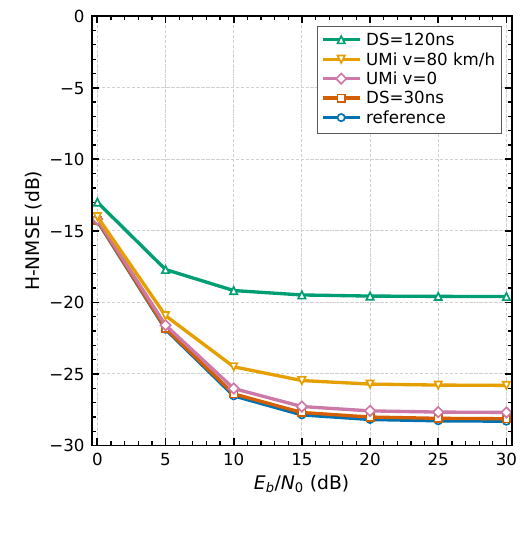}
        \caption{RMS DS and velocity: H-NMSE.}
        \label{fig:gen_ds_velocity_h}
    \end{subfigure}
    \hfill
    \begin{subfigure}[t]{0.32\textwidth}
        \centering
        \includegraphics[width=\linewidth]
            {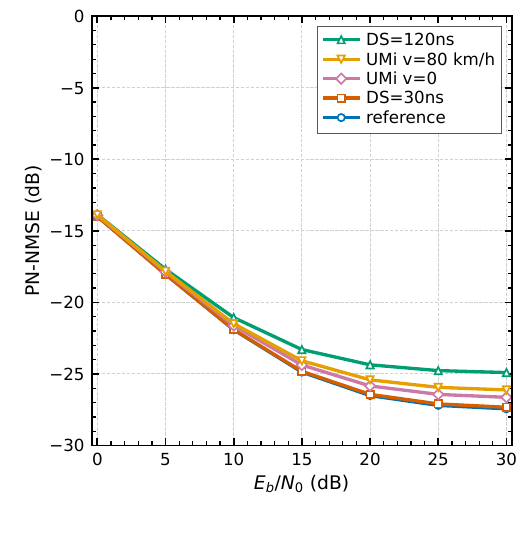}
        \caption{RMS DS and velocity: PN-NMSE.}
        \label{fig:gen_ds_velocity_pn}
    \end{subfigure}
    \hfill
    \begin{subfigure}[t]{0.32\textwidth}
        \centering
        \includegraphics[width=\linewidth]
            {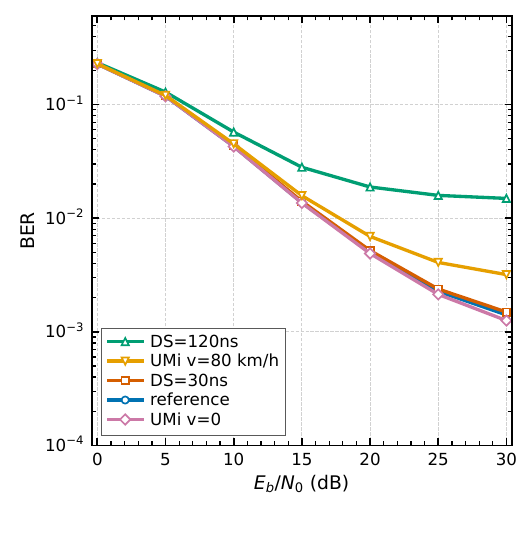}
        \caption{RMS DS and velocity: BER.}
        \label{fig:gen_ds_velocity_ber}
    \end{subfigure}

    \par\smallskip

    \begin{subfigure}[t]{0.32\textwidth}
        \centering
        \includegraphics[width=\linewidth]
            {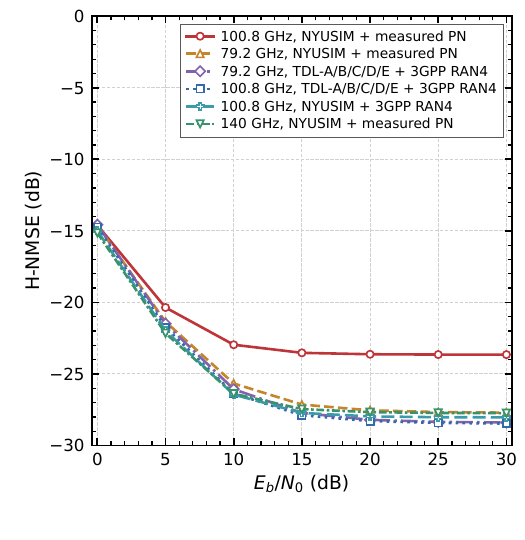}
        \caption{Channel and PN models: H-NMSE.}
        \label{fig:gen_channel_pn_h}
    \end{subfigure}
    \hfill
    \begin{subfigure}[t]{0.32\textwidth}
        \centering
        \includegraphics[width=\linewidth]
            {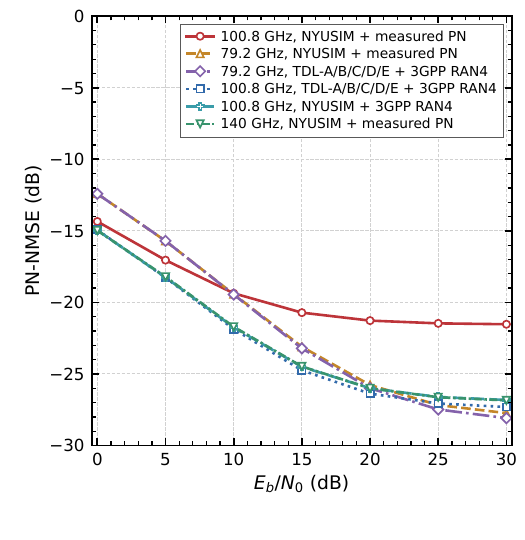}
        \caption{Channel and PN models: PN-NMSE.}
        \label{fig:gen_channel_pn_pn}
    \end{subfigure}
    \hfill
    \begin{subfigure}[t]{0.32\textwidth}
        \centering
        \includegraphics[width=\linewidth]
            {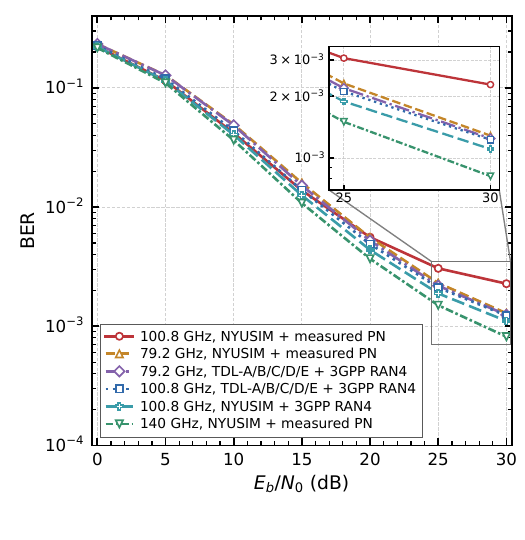}
        \caption{Channel and PN models: BER.}
        \label{fig:generalization_ber}
    \end{subfigure}

    \caption{Generalization and stress tests of the frozen CRFCAN checkpoint without fine-tuning. Each $E_b/N_0$ point uses 2,000 frames per condition.}
    \label{fig:operating_generalization}
\end{figure*}

\begin{figure*}[!t]
    \centering
    \subfloat[Network structure and pilot spacing.\label{fig:ablation_structure}]{%
        \includegraphics[width=0.32\textwidth]
        {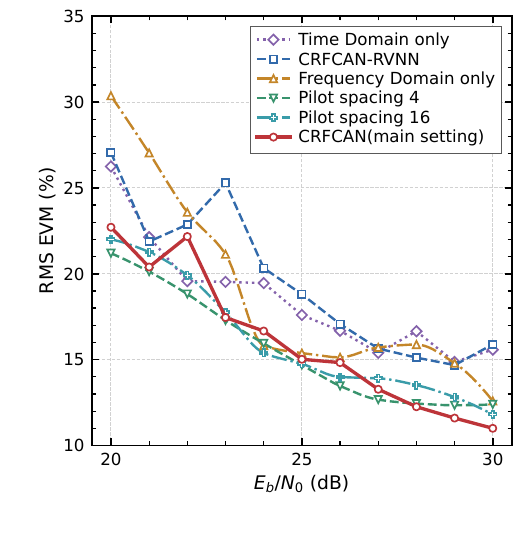}%
    }
    \hfill
    \subfloat[PRCAB placement and PN tail.\label{fig:ablation_modules}]{%
        \includegraphics[width=0.32\textwidth]
        {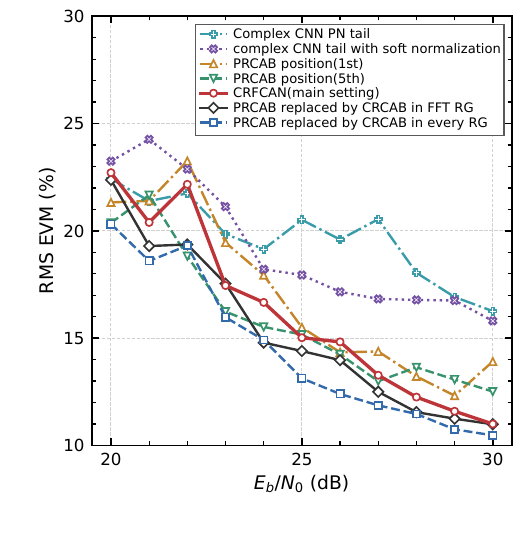}%
    }
    \hfill
    \subfloat[Network depth and width.\label{fig:ablation_scale}]{%
        \includegraphics[width=0.32\textwidth]
        {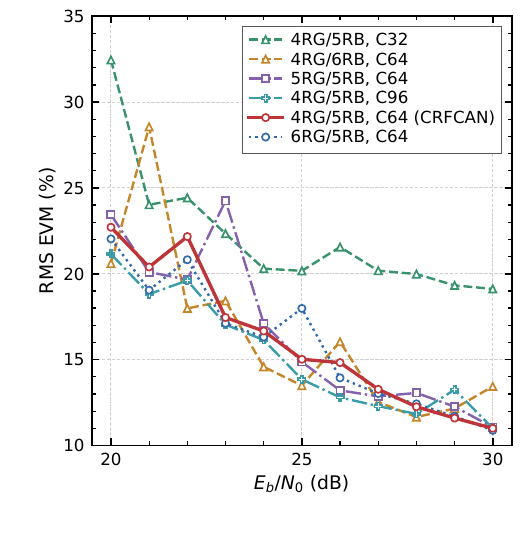}%
    }
    \caption{Receiver EVM for the ablation configurations. Each $E_b/N_0$ point uses 2,000 test frames.}
    \label{fig:ablation_evm}
\end{figure*}

\subsection{Generalization Performance}
\label{sec:generalization}

All tests use the same trained CRFCAN checkpoint as
the preceding benchmark evaluation, without fine-tuning.
Fig.~\ref{fig:operating_generalization}(a)--(c) examines
data modulation, subcarrier spacing, and FFT length
using an equal mixture of InH and UMi scenarios,
with equally represented TDL-A/B/C/D/E profiles within
each scenario and the RAN4 PN model.
Replacing 16-QAM data with QPSK or 64-QAM has little
effect on high-SNR H-NMSE and PN-NMSE, whereas BER
increases with modulation order.
Changing the FFT length from 2048 to 1024 at
480-kHz subcarrier spacing causes only a modest
performance loss.
With FFT1024, further changing the subcarrier spacing
produces different effects: the 240-kHz configuration
exhibits a larger PN estimation error, whereas the
960-kHz configuration shows more pronounced channel
estimation degradation.
At 240~kHz, the longer OFDM symbols experience PN
variation over a longer interval.
At 960~kHz, the shorter CP leaves some previously
admissible channels with insufficient CP coverage.
These results show that CRFCAN can accommodate the
tested OFDM configurations without retraining,
while performance remains affected by the associated
PN variation and multipath interference.

Fig.~\ref{fig:operating_generalization}(d)--(f) examines
delay spread and mobility, including delay-spread stress
cases with insufficient CP coverage, using UMi channels
with equally represented TDL-A/B/C/D/E profiles and the RAN4
PN model for all curves.
The \emph{reference} uses the UMi LOS/NLOS-dependent
RMS delay-spread distributions, conditioned on
$\sigma_\tau\leq100$~ns and sufficient CP coverage,
together with the training mobility mixture of static,
pedestrian, urban-vehicle, and highway-vehicle conditions
with probabilities of $50\%$, $30\%$, $17\%$, and $3\%$,
respectively.
For the \textit{DS=30ns} and \textit{DS=120ns} conditions, the RMS delay-spread parameter is fixed at 30 and 120~ns, respectively, with the same velocity settings as the \textit{reference}.
The path delays are rescaled without reapplying the CP-coverage criterion.
Consequently, 20.8\% of the 30-ns realizations have insufficient CP coverage, yet this condition performs close to the \textit{reference}; all 120-ns realizations have insufficient CP coverage, and channel estimation and BER degrade substantially.
For the mobility comparison, the delay-spread settings follow the \textit{reference}, while the speed is fixed at 80~km/h (\textit{UMi v=80 km/h}) or zero (\textit{UMi v=0}).
Compared with \emph{UMi v=0}, the \emph{UMi v=80 km/h}
condition exhibits a larger degradation in H-NMSE than
in PN-NMSE and approximately 2.5 times the BER at 30~dB.
These results show that CRFCAN tolerates moderate
delay-spread changes, while substantial CP insufficiency
and faster channel variation degrade its estimation performance.

Fig.~\ref{fig:operating_generalization}(g)--(i) examines
changes in channel family, PN spectrum, and carrier
frequency using NYUSIM~\cite{poddar2024nyusim} and PN
synthesized from the measured oscillator spectra at
79.2, 100.8, and 140~GHz reported
in~\cite{plouchart2013synthesizer,
liu2020synthesizer,callender2022transmitter}, respectively.
These spectra are shown in Fig.~\ref{fig:pn_psd}.
The CRFCAN model trained on TDL-A/B/E channels with
RAN4 PN is evaluated without fine-tuning.
Both TDL and NYUSIM channels are set to static, while PN
remains time-varying.
At 100.8~GHz, with the same OFDM configuration and
RAN4 PN model, CRFCAN achieves similar H-NMSE, PN-NMSE,
and BER on TDL and NYUSIM channels.
At $E_b/N_0=30$~dB, their BERs are
$1.23\times10^{-3}$ and $1.11\times10^{-3}$,
respectively.
On the same NYUSIM channels, replacing RAN4 with PN
synthesized from the measured $100.8$-GHz PSD increases
the BER to $2.28\times10^{-3}$ and raises the
H-NMSE and PN-NMSE floors.
In contrast, at $79.2$~GHz, CRFCAN achieves similar
performance with TDL/RAN4 and NYUSIM/measured-PSD PN,
with BERs of $1.25\times10^{-3}$ and
$1.29\times10^{-3}$, respectively, at the same
$E_b/N_0$.
The impact therefore depends on the particular PN
spectrum, with the measured 100.8-GHz PSD presenting
a greater challenge.
The 140-GHz configuration achieves a lower BER.
Owing to the NYUSIM bandwidth limit, the 79.2-GHz
cases use FFT1024 rather than FFT2048, while maintaining
the same subcarrier spacing and CP duration.
The cross-frequency results therefore demonstrate
applicability across the tested channel and oscillator
configurations, rather than a performance gain
attributable to increasing carrier frequency alone.

\subsection{Architecture and Module Ablations}
\label{sec:architecture_ablation}

The main CRFCAN configuration uses four residual groups
with five blocks per group and 64 complex feature channels,
including one PRCAB per group, as shown in
Fig.~\ref{fig:CRFCAN_overview}.
Fig.~\ref{fig:ablation_evm} reports the receiver EVM of the ablation configurations, with all configurations evaluated on the same frames at each $E_b/N_0$ point.
The non-monotonic EVM behavior arises from some frames in which near-zero channel estimates on individual subcarriers amplify equalization errors, producing
EVM outliers for several ablation configurations at
particular SNR points.

Fig.~\ref{fig:ablation_evm}(a) compares complex-valued and
real-valued parameterizations, cross-domain processing,
and pilot spacing.
The real-valued neural network (RVNN) counterpart, \emph{CRFCAN-RVNN}, retains the same overall architecture and block
and tail designs, but uses a reduced feature width to
compensate for the higher parameter count of real-valued
I/Q convolutions.
The main configuration and the approximately
parameter-matched \emph{CRFCAN-RVNN} contain 2.957M and 2.985M parameters,
respectively, differing by less than 1\%.
The complex-valued configuration achieves lower EVM at
all SNRs shown.
The \emph{Frequency Domain only} and \emph{Time Domain only} variants remove the
FFT/IFFT transitions between residual groups while
retaining the other network settings and the required
input/output FFT/IFFT.
All residual groups therefore process features exclusively
in the frequency or time domain, respectively.
The cross-domain main configuration outperforms both
variants, supporting the joint use of these representations.
The pilot-spacing comparison retains the same
rotating-comb structure and varies the spacing from
8 to 4 or 16.
These changes produce SNR-dependent effects without
consistent improvement over spacing 8, indicating that
additional pilots do not necessarily reduce receiver EVM.

Fig.~\ref{fig:ablation_evm}(b) compares PRCAB configurations
and PN estimation tails.
Replacing all PRCABs with CRCABs lowers EVM but increases
the parameter count to 3.484M, approximately 17.8\% above
the main configuration.
The main configuration achieves similar high-SNR
performance with fewer parameters.
Replacing PRCABs with CRCABs only in the FFT residual
groups gives performance close to the main configuration.
The placement comparison retains the main architecture
and changes PRCAB positions only within the FFT residual
groups.
Middle placement gives better medium-to-high-SNR
performance than placement at the first (1st) or last
(5th) block.
The \emph{Complex CNN PN tail} variant replaces the
unfolded PN refinement tail with a CNN regression
structure similar to the channel tail.
Its substantially higher EVM at medium-to-high SNR
supports combining pilot constraints with feature-guided
corrections for PN refinement.
Adding soft normalization to this CNN output, as shown
by \emph{Complex CNN PN tail with soft normalization},
reduces EVM relative to the CNN tail without
output normalization.
The main CRFCAN configuration achieves substantially
lower EVM than both CNN-tail variants, demonstrating
the benefit of the unfolded PN refinement tail.

Fig.~\ref{fig:ablation_evm}(c) examines network width
and depth using \emph{4RG/5RB, C64 (CRFCAN)}, the main CRFCAN configuration, as the
reference.
The \emph{4RG/5RB, C32} variant reduces the parameter
count from 2.957M to 0.824M but produces substantially
higher EVM throughout the plotted SNR range.
Increasing the width to 96 channels
(\emph{4RG/5RB, C96}) raises the parameter count to
6.495M without a consistent high-SNR improvement.
At 64 channels, the \emph{4RG/6RB, C64},
\emph{5RG/5RB, C64}, and \emph{6RG/5RB, C64}
configurations show SNR-dependent differences but
no consistent EVM advantage over the main configuration.
The 4RG/6RB and 6RG/5RB variants increase the parameter
count to 3.552M and 4.330M, respectively.

\subsection{Computational Complexity and Inference Cost}
\label{subsec:complexity_cost}

Table~\ref{tab:benchmark_inference_cost} compares computational
cost, median per-frame inference latency, and peak allocated
GPU memory for
$N_c=2048$, $N_t=8$, and a batch size of one.
All receivers use FP32/complex64 arithmetic.
Neural receivers run on an NVIDIA H100 80\,GB GPU,
while classical receivers run on an AMD EPYC 9654 CPU
with two threads.
FLOPs are analytical real-valued-equivalent operation
counts, including FFT and LS computations, with one
real multiply--accumulate counted as two operations.
For Iterative LS, the reported count is the median
across evaluated frames because its iteration count
is adaptive.
Each receiver retains its native pilot pattern.
Latency excludes data generation, model loading,
and host--device transfers.

\begin{table}[t]
    \centering
    \caption{Receiver computational cost, latency, and peak GPU memory.}
    \label{tab:benchmark_inference_cost}
    \footnotesize
    \setlength{\tabcolsep}{2.5pt}
    \begin{tabular}{lrrrrl}
        \hline
        Method & \shortstack{Params\\(M)} & GFLOPs & \shortstack{Latency\\(ms)} & \shortstack{GPU memory\\(MB)} & Device \\
        \hline
        CRFCAN       & 2.957 & 191.152 &  28.79 & 139.32 & H100 \\
        6RG/5RB, C64 & 4.330 & 279.732 &  37.97 & 161.62 & H100 \\
        4RG/6RB, C64 & 3.552 & 229.967 &  32.40 & 141.72 & H100 \\
        4RG/5RB, C32 & 0.824 &  53.603 &  28.58 &  93.36 & H100 \\
        4RG/5RB, C96 & 6.495 & 419.098 &  31.05 & 199.61 & H100 \\
        \hline
        DeepCascade  & 1.347 &  38.260 &  21.62 & 723.06 & H100 \\
        DeepSRX      & 0.676 &  22.121 &   4.47 &  71.07 & H100 \\
        Iterative LS & --    &   0.794 & 181.59 &     -- & CPU  \\
        PTRS-CPE     & --    & 0.00262 &   3.70 &     -- & CPU  \\
        PTRS-block   & --    & 0.00532 &   4.96 &     -- & CPU  \\
        \hline
    \end{tabular}
    \par\vspace{2pt}
    \parbox{\linewidth}{\scriptsize
        CRFCAN uses 4RG/5RB with C64.
        Scale variants retain the same block and tail designs.
        GPU memory is the peak allocated memory recorded by PyTorch
        ($1~\mathrm{MB}=10^6$ bytes).
    }
\end{table}

Together with Fig.~\ref{fig:ablation_evm}(c), these results
show that increasing the group count, block count, or
feature width raises computational cost without
consistent EVM improvement.
Reducing the width to 32 channels substantially lowers the
parameter count and FLOPs, but increases EVM.
The 64-channel 4RG/5RB configuration therefore offers
a practical performance--cost compromise among the
tested variants.
For 4RG/5RB, increasing the feature width from 32 to
96 channels changes the per-frame latency only from $28.58$ to
$31.05$~ms, whereas the 64-channel 6RG/5RB variant
takes $37.97$~ms.
Thus, in the tested single-frame inference setting,
latency is more sensitive to the number of residual
groups than to feature width.
CRFCAN's channel-estimation and BER gains over the
learned baselines come with higher computational cost
and longer GPU inference latency.
Its peak GPU memory of 139~MB lies between those of
DeepSRX (71~MB) and DeepCascade (723~MB).
The classical receivers require substantially fewer
FLOPs, with PTRS providing the lowest-cost alternatives.

\section{Conclusion}

This paper presented CRFCAN for joint channel and PN
estimation in sub-THz SISO CP-OFDM systems using
comb-type pilots only.
The receiver combines a shared complex-valued cross-domain
backbone, incorporating CRCABs and PRCABs, with an unfolded
PN refinement tail.
In both the training and held-out 3GPP channel profiles and PN spectra, CRFCAN achieves lower channel
estimation error and BER than the evaluated baselines,
with PN estimation gains at medium-to-high SNR.
Further tests without fine-tuning demonstrate its applicability across
the evaluated modulation, OFDM, channel, and PN
configurations, while revealing limitations under
faster channel variation, insufficient CP coverage,
and certain oscillator spectra.
Ablation studies clarify the effects of individual
modules and architectural choices, while runtime and
memory measurements quantify the cost of the proposed
receiver relative to the evaluated baselines.
Future work will extend CRFCAN to joint channel and PN
estimation in MIMO systems, considering spatial pilot
design and shared or independent oscillators across
RF chains.

\bibliographystyle{IEEEtran}
\bibliography{reference}

\end{document}